%% file: main.tex
\documentclass[10pt,twocolumn,letterpaper]{article}
\usepackage[pagenumbers]{cvpr}

\input{preamble}
\usepackage{rotating}
\usepackage{float}
\usepackage{afterpage}
\makeatletter\newcommand\captionof[1]{\def\@captype{#1}\caption}\makeatother
\newcommand{\spos}[1]{\textcolor{posgreen}{\textbf{+#1}}}
\newcommand{\sneg}[1]{#1}
\newcommand{\szero}[1]{#1}
\newtheorem{lemma}{Lemma}[section]
\newtheorem{prop}[lemma]{Proposition}
\newtheorem{corollary}[lemma]{Corollary}
\newtheorem*{propone}{Proposition 1 (restated)}
\newcommand{\R}{\mathbb{R}}
\newcommand{\tr}{\operatorname{tr}}
\newcommand{\range}{\operatorname{range}}
\definecolor{cvprblue}{rgb}{0.21,0.49,0.74}
\usepackage[pagebackref,breaklinks,colorlinks,allcolors=cvprblue]{hyperref}

\title{Merge++: Universal Merge Refinement Through Data-Free Checkpoint Inversion}
\author{%
Aditya Pola\textsuperscript{1}\hspace{3em}Vineeth N. Balasubramanian\textsuperscript{1,2}\\[0.6em]
{\normalsize\textsuperscript{1}IIT Hyderabad\hspace{3em}\textsuperscript{2}Microsoft Research, India}%
}

\begin{document}
\crefname{prop}{Proposition}{Propositions}\Crefname{prop}{Proposition}{Propositions}
\crefname{lemma}{Lemma}{Lemmas}\Crefname{lemma}{Lemma}{Lemmas}
\crefname{corollary}{Corollary}{Corollaries}\Crefname{corollary}{Corollary}{Corollaries}
\maketitle
\addtocontents{toc}{\protect\setcounter{tocdepth}{-10}}
\input{sec/0_abstract}
\input{sec/1_intro}
\input{sec/2_related}
\input{sec/arxiv_3_method}
\input{sec/arxiv_4_experiments}
\input{sec/5_conclusion}
\input{sec/9_llm_statement}
{
    \small
    \bibliographystyle{ieeenat_fullname}
    \bibliography{main}
}

\clearpage
\onecolumn
\appendix
\numberwithin{equation}{section}
\numberwithin{table}{section}
\numberwithin{figure}{section}
\setlength{\abovedisplayskip}{9pt plus 2pt}
\setlength{\belowdisplayskip}{9pt plus 2pt}
\begin{center}{\Large\textbf{Appendix}}\end{center}
\addtocontents{toc}{\protect\setcounter{tocdepth}{2}}
\tableofcontents
\input{sec/A_appendix}
\end{document}

%% file: preamble.tex
\usepackage{colortbl}
\definecolor{posgreen}{rgb}{0.0, 0.5, 0.0}
\definecolor{negred}{rgb}{0.7, 0.0, 0.0}
\definecolor{grayrow}{gray}{0.93}
\definecolor{tickgray}{gray}{0.88}
\definecolor{altrow}{gray}{0.94}
\newcommand{\pos}[1]{\textcolor{posgreen}{\small\textbf{+#1}}}

\newcommand{\zerov}[1]{{\small\textbf{#1}}}

\usepackage{amsthm}
\newtheorem{proposition}{Proposition}

%% file: sec/0_abstract.tex
\begin{abstract}
Model merging consolidates fine-tuned experts into one multi-task model without retraining. All existing data-free methods approach this problem entirely in weight space. Restricted to arithmetic on parameters, these methods never observe how each expert behaves, a signal that only emerges through forward evaluation. Accessing this behavioural signal requires inputs to evaluate on, which the data-free setting prohibits. We propose Merge++, a post-hoc method that inverts the expert checkpoints to synthesise task-representative images and then uses these images to distill the experts into the merged model. Because all experts are distilled into the same weights, the updates induced by the images of different tasks can interfere with one another. To avoid such interference, we constrain the images of each task to the part of its task vector's subspace that is not shared with other tasks. Merge++ requires no data beyond the checkpoints and can be applied on top of any weight-space merging algorithm. It consistently improves merging algorithms ranging from simple task arithmetic to state-of-the-art spectral and optimisation-based methods across three architectures and three task scales, with average gains of up to 6.6 points.
\end{abstract}

%% file: sec/1_intro.tex
\section{Introduction}

Fine-tuning pretrained models on downstream tasks has become standard practice in both vision and language, producing a growing number of task-specific checkpoints hosted on open repositories. However, deploying a separate model for each task incurs storage and inference costs that scale linearly with the number of tasks. Multi-task learning addresses this by training a single model jointly, but requires simultaneous access to all training data, which is often impractical due to privacy or licensing constraints. Model merging offers an alternative by combining the parameters of independently fine-tuned models into one multi-task model without any additional training. The only requirement is that the models share a common pretrained initialisation. This has made model merging an active area of research, centred on how to combine the parameters of multiple experts into a single model that preserves the performance on each task.

\begin{figure*}[t!]
    \centering
    \includegraphics[width=\textwidth]{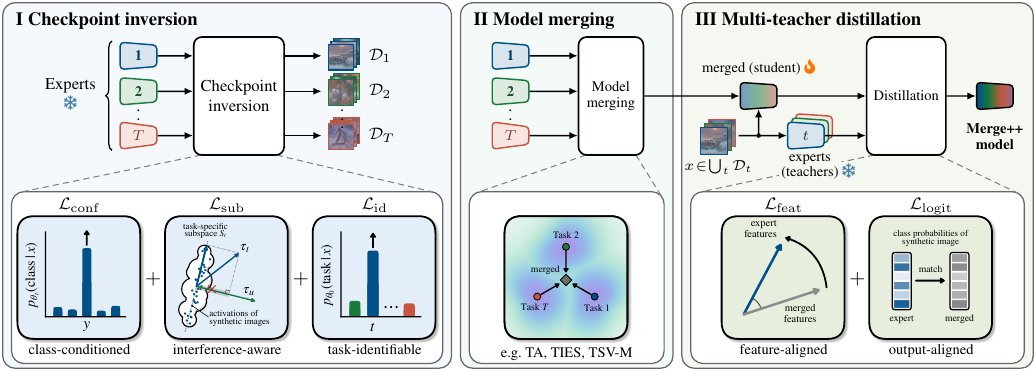}
    \caption{\textbf{Overview of Merge++.} \textbf{(I) Checkpoint inversion.} Each expert is inverted to synthesise images of its task without real data. The synthesis is (a) class-conditioned, maximising the expert's probability of each image's target class, (b) interference-aware, concentrating the activations of each task's images in a task-specific subspace, and (c) task-identifiable, maximising the probability that the pretrained model assigns each image to its task. \textbf{(II) Model merging.} The experts are merged with any weight-space method. \textbf{(III) Multi-teacher distillation.} The merged model is trained to match each expert on the synthetic images of its task, yielding the Merge++ model.}
    \label{fig:top_full_width}
\end{figure*}

The most practical merging setting assumes no access to training data, with only the checkpoints available. This data-free constraint rules out methods that require validation sets for tuning or test-time adaptation. Within this setting, existing methods define a \emph{task vector} as the difference between each fine-tuned model and the pretrained initialisation, and reduce merging to combining these task vectors into a single update, by scaling and summing them~\citep{ilharco2023task}, resolving conflicts between their parameters~\citep{yadav2023ties}, separating their singular subspaces~\citep{tang2025tsvm,lu2025isomerging}, or optimising the merge against an interference objective~\citep{cheng2025wudi}. These methods have steadily narrowed the gap to the individual experts, but a substantial gap remains, and it widens as more tasks are merged. Despite their diversity, all of these methods share a common trait in that they operate entirely on parameter tensors, with no forward computation to reveal how each expert behaves on inputs. Methods that do observe this behaviour, by adapting the merge on unlabelled test data~\citep{yang2024adamerging} or distilling the experts on a few labelled examples~\citep{xu2025prodistill}, give up the data-free setting.

To observe the behaviour of the experts without real data, we turn to model inversion. Prior work on data-free knowledge transfer has shown that a trained network can be inverted to synthesise images representative of its training distribution, by optimising random noise to maximise the network's confidence on target classes \citep{yin2020dreaming}. For vision transformers, which store no batch-normalisation statistics to guide this process, realistic images can instead be obtained with a multi-scale image prior and augmentations \citep{cazenavette2025lgm}. The fine-tuned experts are natural candidates for inversion, as each encodes the structure of its task. Synthesising images from every expert therefore provides the inputs that data-free merging lacks, and distilling the experts on these images into the merged model corrects its behaviour on each task without real data. Since the images depend only on the checkpoints, they are synthesised once and can refine the output of any merging method, making this a complementary stage rather than a new merging algorithm.

Distilling all experts into a single model, however, raises a challenge that does not arise in single-model distillation. Every synthetic image updates the same shared weights, and the updates induced by the images of different tasks can interfere with one another, just as task vectors do in weight space. We propose Merge++, a post-hoc refinement method that synthesises images designed to avoid this interference (Figure~\ref{fig:top_full_width}). For each task, Merge++ extracts from the task vectors a subspace that is not shared with the other tasks. In addition to standard inversion, the synthetic images of that task are optimised to activate mainly this subspace and to be distinguishable from the images of other tasks by the pretrained model. The merged model is then refined by distilling each expert on its own synthetic images.

Our contributions are threefold. \textbf{(1)}~We introduce Merge++, a post-hoc refinement that complements existing weight-space merging methods by distilling the experts into the merged model using images synthesised from their checkpoints. \textbf{(2)}~We show that the updates induced by the synthetic images of different tasks can interfere, and propose a synthesis objective that separates the tasks in activation space, using subspaces derived from the task vectors. \textbf{(3)}~We show that Merge++ consistently improves state-of-the-art merging methods, supported by ablations and an analysis of where refinement acts in the network.

%% file: sec/2_related.tex
\section{Related Work}

\paragraph{Model merging.} Model merging combines experts fine-tuned from a common pretrained model into a single model without retraining. Task Arithmetic~\citep{ilharco2023task} showed that experts can be combined by adding their task vectors, but the merged model falls well short of the individual experts because the task vectors interfere. Subsequent work has largely sought better ways to resolve this interference in weight space. Early methods treat parameters individually, trimming small updates and resolving sign conflicts~\citep{yadav2023ties}, randomly dropping and rescaling updates~\citep{yu2024dare}, or removing weights relevant to too few tasks~\citep{wang2024tall}. Spectral methods exploit the structure of each layer's task matrix instead, decorrelating the singular vectors of different tasks~\citep{tang2025tsvm} or separating a common subspace from task-specific ones~\citep{lu2025isomerging}. Most recently, WUDI~\citep{cheng2025wudi} and SWUDI-A~\citep{wei2026swudi} optimise the merged weights against a proxy of interference derived from the task vectors. Despite their increasing sophistication, all of these methods define and measure interference on the parameters alone, and never observe how the merged model behaves on inputs. Methods that do observe this behaviour, by learning merging coefficients on unlabelled test data~\citep{yang2024adamerging}, correcting the merged representations~\citep{yang2024surgery}, or distilling the experts layer by layer~\citep{xu2025prodistill}, require access to real data. Merge++ obtains this behavioural signal without data, by synthesising inputs from the checkpoints themselves.

\paragraph{Data-free knowledge transfer.} When a model's training data is unavailable, its knowledge can still be transferred by synthesising inputs from the model itself. Generator-based methods train a generator to produce inputs on which the teacher is confident or the student disagrees~\citep{chen2019dafl, micaelli2019zskt, do2022mad}, whereas inversion-based methods optimise the inputs directly. DeepInversion~\citep{yin2020dreaming} matches the batch-normalisation statistics stored in the teacher, and subsequent work improves the diversity~\citep{fang2021cmi} and efficiency~\citep{fang2022fast, tran2024nayer} of the synthesis. Vision transformers store no such statistics, so their inversion relies instead on image priors and augmentations~\citep{olah2017feature, ghiasi2022plugin, hu2024sparse}. Knowledge amalgamation extends distillation to several teachers, training a single student to reproduce all of them~\citep{shen2019amalgamating, ye2019student}, including without data~\citep{ye2020datafree} and in federated learning~\citep{zhang2022dense, zhang2022fedftg, dai2024coboosting}. In all of these settings, the synthetic inputs are designed for each teacher in isolation, and the student does not share its parameters with the teachers. Merge++ differs in that the student is a merge of the teachers, so the synthetic images of different teachers update the same shared weights and can interfere with one another. To our knowledge, ours is the first method that accounts for this interference when synthesising the data, by concentrating the activations of each task in a subspace derived from the task vectors.

\paragraph{Dataset condensation.} Dataset condensation compresses a large training set into a small set of synthetic images on which a model can be trained to comparable accuracy. The synthetic images are optimised to reproduce the effect of the real data on training, by matching the gradients~\citep{zhao2021dc}, feature distributions~\citep{zhao2023dm}, or training trajectories~\citep{cazenavette2022mtt} that it induces, and subsequent work scales this to large datasets~\citep{yin2023sre2l}. Most relevant to us, linear gradient matching~\citep{cazenavette2025lgm} condenses datasets for linear probes on pretrained vision backbones, and shows that a multi-scale image parameterisation yields realistic synthetic images that transfer across models. We adopt this parameterisation. Unlike condensation, however, our synthesis has no access to real data: the images are obtained from the expert checkpoints alone, and serve to refine a merged model rather than to train a model from scratch.

%% file: sec/arxiv_3_method.tex
\section{Method}

\subsection{Problem Setup}

Let $\theta_0$ denote the parameters of a pretrained model and $\{\theta_t\}_{t=1}^T$ denote $T$ models independently fine-tuned on separate tasks from the same initialisation. The task vector for task $t$ is defined as $\tau_t = \theta_t - \theta_0$. Let $f_\theta$ denote the image encoder with parameters $\theta$, and let $W_t$ denote the classification head for task $t$, whose rows are the CLIP text embeddings of the task's classes $C_t$~\citep{radford2021clip}. The heads remain fixed during fine-tuning and are shared by the pretrained, fine-tuned, and merged encoders, as is standard in model merging. Data-free model merging seeks a single parameter vector $\theta_m$ that performs well across all $T$ tasks, using only the checkpoints $\theta_0, \theta_1, \ldots, \theta_T$ without access to their training data. Existing methods produce $\theta_m$ by combining the task vectors in parameter space, for example $\theta_m = \theta_0 + f(\tau_1, \ldots, \tau_T)$ where $f$ varies across methods. Merge++ takes any such $\theta_m$ and produces a refined $\theta_m^+$ in two stages, described below.

\subsection{Interference-Aware Expert Inversion}
\label{sec:inversion}

For each expert $\theta_t$, we synthesise a set $\mathcal{D}_t$ of $M$ images by optimising them from random noise against the expert. These images capture the task knowledge encoded in $\theta_t$. Since they are used to refine the merged model, they are further constrained to improve performance on task $t$ without degrading performance on the remaining tasks.

\paragraph{Image parameterisation.} Following \citet{fort2025das} and \citet{cazenavette2025lgm}, we represent each image as a sum of learnable components at multiple resolutions rather than as raw pixels, since directly optimising pixels against a network yields high-frequency, adversarial patterns~\citep{olah2017feature}. The components form a pyramid $P=\{P_r \mid r\in\rho\}$ with resolutions $\rho=\{1{\times}1,\,2{\times}2,\,4{\times}4,\dots,128{\times}128,\,224{\times}224\}$, and the image is rendered by bilinearly upsampling every level to $224{\times}224$ and summing the results:
\begin{equation}
    x = g(P) = \sigma\Big(C\sum_{r\in\rho}\mathrm{resize}_{224}(P_r)\Big),
    \label{eq:render}
\end{equation}
where $C$ is a fixed linear transform that maps the optimised, decorrelated colour channels to the standard colour space~\citep{olah2017feature}, and the sigmoid $\sigma$ bounds pixel values to $[0,1]$. Optimisation starts from the $1{\times}1$ level alone and adds the next finer level at regular intervals, so that the coarse layout of an image is determined before its fine detail. At every step, we further apply differentiable augmentations~\citep{zhao2021dsa} to $K$ independent copies of each image (random horizontal flipping, random resized cropping, and additive Gaussian noise) and average the loss over the copies.

\paragraph{Confidence term.} As in standard model inversion~\citep{yin2020dreaming}, we optimise each image so that the expert $\theta_t$ classifies it as its assigned target class $y \in C_t$, by minimising the cross-entropy of the expert's prediction on an augmented copy $\tilde{x}$ of the image:
\begin{equation}
    \mathcal{L}_{\mathrm{conf}} = \mathrm{CE}\big(W_t f_{\theta_t}(\tilde{x}),\, y\big).
    \label{eq:conf}
\end{equation}
However, $\mathcal{L}_{\mathrm{conf}}$ only requires that the expert recognises the image. It does not consider that the image is later used to train the merged model, which performs all tasks with one set of parameters. A training step on an image of task $t$ therefore also affects the predictions of the merged model on the other tasks. Such an update can degrade the performance of the merged model on those tasks, or partly reverse the corrections obtained from their own synthetic images.

\paragraph{Task-specific subspaces.} The refinement stage trains the merged model on the synthesised images by gradient descent. For any linear layer, the gradient with respect to its weight matrix takes the form $\sum \delta a^\top$, where $a$ denotes the input activations of the layer and $\delta$ the gradients at its output. Consequently, an image changes the weights of a linear layer only along the directions of the input it provides to that layer. When images of different tasks produce inputs along the same directions, their updates act on the same components of the weight matrix and interfere with one another. To prevent this, we identify for each task and each layer a set of input directions specific to that task, and later encourage its images to produce activations along them.

We obtain these directions from the task vectors alone. Let $\mathcal{L}$ denote the query--key--value projection and the two MLP linear layers of every transformer block, which together hold about 90\% of the encoder's parameters. For a layer $\ell \in \mathcal{L}$, let $\tau_t^\ell$ denote the component of the task vector $\tau_t$ associated with its weight matrix, and let $V_t^\ell$ be the matrix whose columns are the top-$k$ right singular vectors of $\tau_t^\ell$, where $Tk$ does not exceed the input dimension of the layer. The columns of $V_t^\ell$ span the input directions along which fine-tuning on task $t$ changed the response of the layer the most. Since other tasks may have changed the layer along some of the same directions, we remove from $V_t^\ell$ its component in the span of the other tasks' directions and re-orthonormalise the result:
\begin{equation}
    S_t^\ell = \mathrm{orth}\big(V_t^\ell - QQ^\top V_t^\ell\big),
    \label{eq:subspace}
\end{equation}
where $Q$ is an orthonormal basis for the columns of $\{V_u^\ell\}_{u \neq t}$. The resulting subspace $S_t^\ell$ is spanned by input directions that fine-tuning on task $t$ changed substantially and that are orthogonal to the dominant directions of every other task. Since the directions of different tasks are linearly independent in practice, $S_t^\ell$ has the same dimension $k$ as $V_t^\ell$, so every task has a $k$-dimensional subspace in which its activations can be concentrated. It is computed once from the checkpoints, for every layer in $\mathcal{L}$.

\paragraph{Subspace term.} We now optimise the images of task $t$ so that their input activations at each layer $\ell \in \mathcal{L}$ lie predominantly in $S_t^\ell$. For a batch of augmented images of task $t$, let $A^\ell$ denote the matrix whose rows are the mean-centred input activations of layer $\ell$ of the pretrained model $\theta_0$, one per token. The share of the activation energy that lies in $S_t^\ell$ is then given by the ratio $\|A^\ell S_t^\ell\|_F^2 / \|A^\ell\|_F^2$. It equals $0$ for activations orthogonal to $S_t^\ell$ and $1$ for activations within it. The subspace term maximises this share across all layers in $\mathcal{L}$:
\begin{equation}
    \mathcal{L}_{\mathrm{sub}} = -\frac{1}{|\mathcal{L}|}\sum_{\ell\in\mathcal{L}} \log \frac{\|A^\ell S_t^\ell\|_F^2}{\|A^\ell\|_F^2}.
    \label{eq:sub}
\end{equation}

\paragraph{Task identifiability term.} The subspace term constrains the activations of each layer separately. We complement it with a constraint on the output of the pretrained model: images of task $t$ should be recognisable as belonging to task $t$ rather than to another task. To this end, we stack the classification heads of all tasks, so that $\theta_0$ classifies an image over the union of their classes, and maximise the probability that $\theta_0$ assigns to the classes of task $t$:
\begin{equation}
    \mathcal{L}_{\mathrm{id}} = -\log \sum_{c \in C_t} \mathrm{softmax}\big([W_1; \dots; W_T]\, f_{\theta_0}(\tilde{x})\big)_c.
    \label{eq:id}
\end{equation}

\paragraph{Overall objective.} The images are optimised under a single loss that combines the three terms:
\begin{equation}
    \mathcal{L} = \mathcal{L}_{\mathrm{conf}} + w_{\mathrm{sub}}\,\mathcal{L}_{\mathrm{sub}} + w_{\mathrm{id}}\,\mathcal{L}_{\mathrm{id}}.
    \label{eq:total}
\end{equation}
The gradients of the three terms with respect to $P$ differ widely in magnitude, and their ratios change as the images evolve. Rather than fixed weights, which would need tuning for every task, we use adaptive weights recomputed at every step, so that each additional term contributes a fixed fraction $\lambda$ of the gradient norm of the confidence term:
\begin{equation}
    w_j = \lambda\, \frac{\|\nabla_P \mathcal{L}_{\mathrm{conf}}\|}{\|\nabla_P \mathcal{L}_j\|}, \qquad j \in \{\mathrm{sub}, \mathrm{id}\}.
    \label{eq:weights}
\end{equation}
The weights are treated as constants when differentiating $\mathcal{L}$. The pyramid $P$ is optimised with Adam, and the rendered images $g(P)$ after the final step form the synthetic set $\mathcal{D}_t$.

\begin{table*}[t]
\centering
\caption{Average top-1 accuracy (\%) across tasks of each base merge and its improvement after applying Merge++.}
\label{tab:main}
\setlength{\tabcolsep}{3.5pt}
\begin{tabular}{ll|ccc|ccc|ccc}
\toprule
& & \multicolumn{3}{c|}{\textbf{ViT-B/32}} & \multicolumn{3}{c|}{\textbf{ViT-B/16}} & \multicolumn{3}{c}{\textbf{ViT-L/14}} \\
\textbf{Method} & & 8T & 14T & 20T & 8T & 14T & 20T & 8T & 14T & 20T \\
\midrule
TA {\tiny\color{gray}[ICLR'23]} & base & 70.0 & 64.9 & 61.2 & 75.6 & 70.6 & 65.8 & 83.9 & 79.3 & 73.8 \\
\rowcolor{grayrow}
& {\footnotesize\textbf{+MERGE++}} & \pos{7.6} & \pos{5.8} & \pos{5.6} & \pos{5.3} & \pos{5.3} & \pos{4.2} & \pos{4.3} & \pos{3.2} & \pos{6.4} \\
\midrule
TIES {\tiny\color{gray}[NeurIPS'23]} & base & 73.3 & 67.4 & 63.1 & 78.8 & 72.9 & 68.0 & 85.5 & 79.2 & 75.5 \\
\rowcolor{grayrow}
& {\footnotesize\textbf{+MERGE++}} & \pos{7.0} & \pos{5.3} & \pos{5.9} & \pos{4.5} & \pos{4.5} & \pos{4.4} & \pos{4.3} & \pos{3.7} & \pos{5.5} \\
\midrule
Iso-CTS {\tiny\color{gray}[ICML'25]} & base & 83.5 & 80.2 & 77.2 & 89.2 & 84.9 & 81.6 & 92.9 & 89.6 & 89.3 \\
\rowcolor{grayrow}
& {\footnotesize\textbf{+MERGE++}} & \pos{2.6} & \pos{2.2} & \pos{3.6} & \pos{0.2} & \pos{0.8} & \pos{2.5} & \pos{0.3} & \pos{1.1} & \pos{1.5} \\
\midrule
TSV-M {\tiny\color{gray}[CVPR'25]} & base & 83.7 & 79.2 & 76.1 & 87.2 & 83.7 & 80.2 & 91.2 & 88.2 & 87.2 \\
\rowcolor{grayrow}
& {\footnotesize\textbf{+MERGE++}} & \pos{2.5} & \pos{2.8} & \pos{5.0} & \pos{1.4} & \pos{1.3} & \pos{3.6} & \pos{1.1} & \pos{1.5} & \pos{2.6} \\
\midrule
WUDI {\tiny\color{gray}[ICML'25]} & base & 84.9 & 77.7 & 62.5 & 88.4 & 82.4 & 69.8 & 92.1 & 88.9 & 82.6 \\
\rowcolor{grayrow}
& {\footnotesize\textbf{+MERGE++}} & \pos{2.2} & \pos{4.7} & \pos{16.5} & \pos{1.0} & \pos{2.5} & \pos{11.4} & \pos{0.6} & \pos{1.3} & \pos{6.6} \\
\midrule
SWUDI-A {\tiny\color{gray}[arXiv'26]} & base & 85.6 & 81.3 & 78.5 & 89.1 & 85.7 & 82.3 & 92.5 & 90.2 & 89.1 \\
\rowcolor{grayrow}
& {\footnotesize\textbf{+MERGE++}} & \pos{1.6} & \pos{1.3} & \pos{3.0} & \pos{0.6} & \zerov{0.0} & \pos{1.7} & \pos{0.4} & \pos{0.4} & \pos{1.2} \\
\midrule
\rowcolor{grayrow}
\multicolumn{2}{l|}{\textbf{Avg.\ $\Delta$}} & \pos{3.9} & \pos{3.7} & \pos{6.6} & \pos{2.2} & \pos{2.4} & \pos{4.6} & \pos{1.8} & \pos{1.9} & \pos{4.0} \\
\bottomrule
\end{tabular}
\end{table*}

\subsection{Multi-Teacher Distillation}

Given a merged model $\theta_m$ produced by a weight-space merging method and the synthetic sets $\mathcal{D}_1, \dots, \mathcal{D}_T$, we refine $\theta_m$ by distilling the knowledge of all experts into it simultaneously. The student $\theta$ is initialised from $\theta_m$, with only the weight matrices of the linear layers in the transformer blocks trainable. For a synthetic image $x$ generated from expert $\theta_t$, we define a feature alignment loss
\begin{equation}
    \mathcal{L}_{\mathrm{feat}} = 1 - \cos\big(f_\theta(x),\, f_{\theta_t}(x)\big),
    \label{eq:feat}
\end{equation}
which encourages the student to produce representations that match those of the expert. We additionally define a logit matching loss~\citep{hinton2015distilling}
\begin{equation}
    \mathcal{L}_{\mathrm{logit}} = \kappa^2\, \mathrm{KL}\big(\mathrm{softmax}(z_t/\kappa)\,\big\|\,\mathrm{softmax}(z/\kappa)\big),
    \label{eq:logit}
\end{equation}
where $z = W_t f_\theta(x)$ and $z_t = W_t f_{\theta_t}(x)$ are the logits of the student and the expert under the classification head of task $t$, and $\kappa$ is a temperature. This transfers the expert's predictive distribution to the student. The student is trained to minimise
\begin{equation}
    \mathcal{L}_{\mathrm{KD}} = \mathbb{E}_{(x,t)}\big[\mathcal{L}_{\mathrm{feat}}(x) + \mathcal{L}_{\mathrm{logit}}(x)\big],
    \label{eq:kd}
\end{equation}
where $x$ is a randomly augmented image drawn from the union of the synthetic sets and $t$ is the index of the expert that generated it. Each image is thus paired with its generating expert as teacher, so the student simultaneously learns to match all experts on their respective tasks. The trained student is the refined model $\theta_m^+$. Since the synthetic sets depend only on the checkpoints and not on the merging method, they are generated once for a given set of tasks, and their cost is amortised over every merge of these experts that is subsequently refined.

\begin{figure*}[t]
\centering
\includegraphics[width=\textwidth]{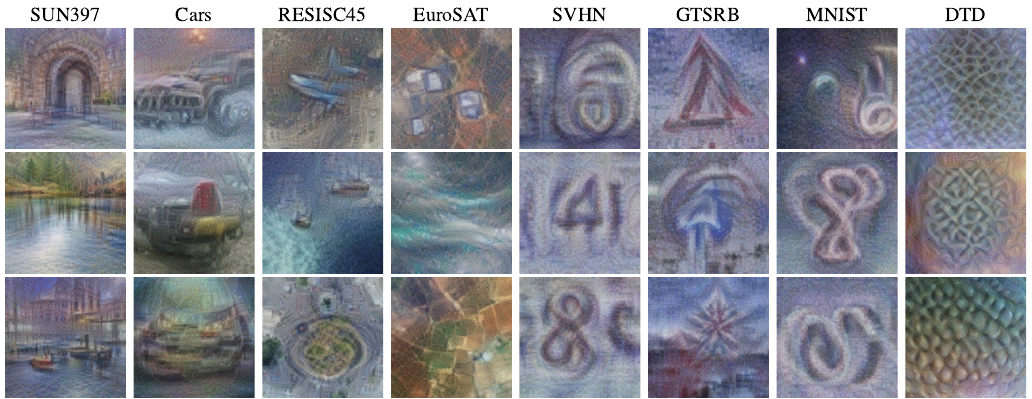}
\caption{Synthetic images generated by Merge++ from the ViT-B/32 experts of the 8-task benchmark, without any real data. Each column shows three images of one task, targeting different classes.}
\label{fig:synth}
\end{figure*}

\subsection{What Can Synthetic Data Correct?}
\label{sec:correct}

Since distillation updates a layer only along the activations of the synthetic images (Sec.~\ref{sec:inversion}), the synthetic data determine which differences between the merge and the experts it can remove. We now make this precise. Following RegMean~\citep{jin2023regmean} and WUDI~\citep{cheng2025wudi}, we consider a single linear layer with fixed inputs, refined by matching the outputs of the experts. This simplification does not describe exactly the effect of distillation on a deep network, but shows which properties of the synthetic data matter for refinement.

Let $\theta_t=\theta_0+\tau_t$ and $\theta_m$ denote the weights of expert $t$ and of the merge at this layer, and let the rows of $A_t$ be the input activations of the synthetic images of task $t$. Refinement minimises
\begin{equation}
    \mathcal{L}(\theta)=\sum_{t=1}^{T}\big\|A_t(\theta-\theta_t)^\top\big\|_F^2
    \label{eq:surrogate}
\end{equation}
by gradient descent from $\theta_m$. For ease of exposition, we assume that the Gram matrices of all tasks are simultaneously diagonalisable, \ie, share an orthonormal eigenbasis $\{s_j\}$, and write
\[
A_t^\top A_t=\sum_j g_{tj}\,s_js_j^\top ,\qquad g_{tj}=\|A_ts_j\|^2 .
\]
The eigenvalue $g_{tj}$ is the \emph{energy} of the activations of task $t$ along $s_j$, \ie, the sum of their squared projections onto $s_j$, and $g_j=\sum_t g_{tj}$ is the total energy along $s_j$. The requirements derived below hold without this assumption, as shown in Appendix~\ref{supp:general}.

\begin{proposition}\label{prop:refine}
Gradient descent on~\eqref{eq:surrogate} converges to a minimiser $\theta^\star$ of $\mathcal{L}$ that satisfies, for every direction $s_j$,
\begin{align}
\theta^\star s_j&=\theta_m s_j &&\text{if } g_j=0, \label{eq:unchanged}\\
\theta^\star s_j&=\sum_{t=1}^{T}\frac{g_{tj}}{g_j}\,\theta_t s_j &&\text{if } g_j>0. \label{eq:average}
\end{align}
The irreducible loss is
\begin{equation}
\mathcal{L}(\theta^\star)=\sum_{j:\,g_j>0}\;\frac{1}{g_j}\sum_{t<u} g_{tj}\,g_{uj}\,\big\|(\tau_t-\tau_u)\,s_j\big\|^2 .
\label{eq:remaining}
\end{equation}
\end{proposition}
The proof is given in Appendix~\ref{supp:prop1}.

According to Eq.~\eqref{eq:unchanged}, refinement leaves the merge unchanged along directions on which no synthetic activation lies, so any error of the merge along these directions persists. The synthetic images of task $t$ should therefore \textbf{(I)} have energy along the directions on which $\tau_t$ acts. Along every other direction, Eq.~\eqref{eq:average} replaces the merge by an average of the experts, weighted by each task's energy. Eq.~\eqref{eq:remaining} gives the irreducible loss, \ie, the mismatch between the refined layer and the experts that no choice of weights removes. A pair of tasks contributes to it along $s_j$ only if both have energy along $s_j$ and their task vectors disagree there ($\tau_t s_j\neq\tau_u s_j$). The activations of the synthetic images of different tasks should therefore \textbf{(II)} not share directions on which their task vectors disagree. The subspace term (Eq.~\eqref{eq:sub}) works towards both by concentrating the activations of task $t$ in $S_t$, which is built from the main directions of $\tau_t$ with those of the other task vectors removed.

%% file: sec/arxiv_4_experiments.tex
\section{Experiments}

\begin{figure*}[t]
\centering
\includegraphics[width=\textwidth]{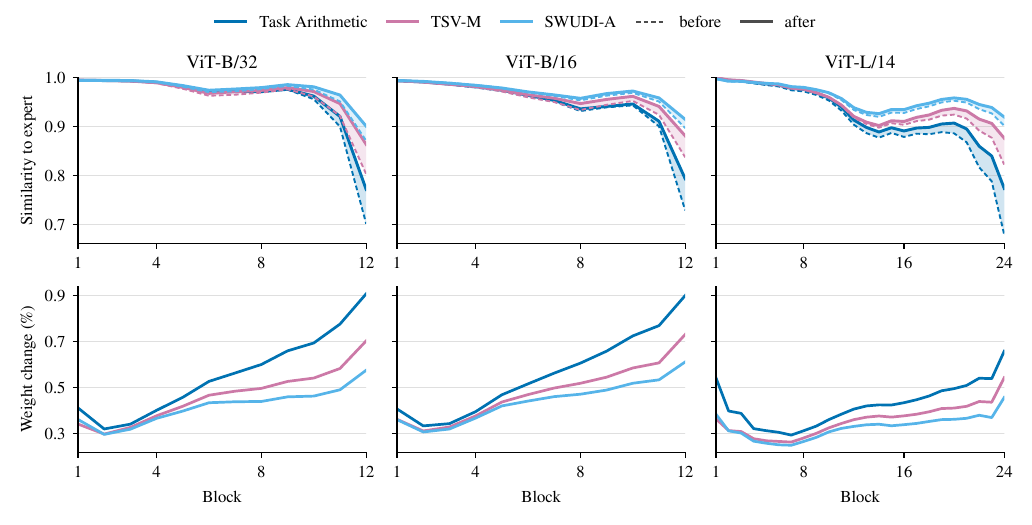}
\caption{Top: cosine similarity between the class-token representations of the merged model and the corresponding expert at the output of each block, on real test images, before (dashed) and after (solid) applying Merge++; the shaded region is the gain. Bottom: relative change of the weights of each block, $\|\Delta W\|/\|W\|$, in percent. Shown for an arithmetic (Task Arithmetic), a spectral (TSV-M), and an optimisation-based (SWUDI-A) base merge; the other three are shown in Appendix~\ref{supp:analysis}.}
\label{fig:analysis}
\end{figure*}

\paragraph{Setup.} We follow the standard evaluation protocol established by \citet{ilharco2023task} and adopted by subsequent work \citep{yadav2023ties, huang2024emr}. We use three CLIP visual encoders as pretrained models: ViT-B/32, ViT-B/16, and ViT-L/14 \citep{radford2021clip}. We evaluate on three task scales \citep{wang2024tall}: 8 tasks (SUN397, Cars, RESISC45, EuroSAT, SVHN, GTSRB, MNIST, DTD), 14 tasks (adding Flowers102, PCAM, FER2013, OxfordPets, STL10, CIFAR100), and 20 tasks (further adding CIFAR10, Food101, FashionMNIST, EMNIST, KMNIST, RenderedSST2). Performance is measured by top-1 accuracy on each task's test set, averaged across all tasks.

\paragraph{Baselines.} We apply Merge++ on top of six data-free merging methods spanning three categories: arithmetic (\citet{ilharco2023task}; \citet{yadav2023ties}), spectral (\citet{lu2025isomerging}; \citet{tang2025tsvm}), and optimisation-based (\citet{cheng2025wudi}; \citet{wei2026swudi}). For each baseline, we report both the original merged accuracy and the accuracy after applying Merge++, along with the improvement. All baselines use their official implementations and hyperparameter selection protocols, and Merge++ is applied to the resulting merges without further tuning. All methods, including Merge++, use the same checkpoints and heads.

\paragraph{Implementation details.} All experts are the fine-tuned CLIP encoders released by \citet{wang2024tall}, together with their classification heads. For the inversion stage, we generate $M=256$ images per task, optimising each pyramid for 450 steps with Adam (learning rate $2\times10^{-3}$) over $K=10$ augmented copies per step, with $k=16$ and $\lambda=0.3$. For the distillation stage, we train for 600 steps with AdamW (learning rate $2\times10^{-6}$), batch size 1024, and temperature $\kappa=2$. All hyperparameters are shared across architectures, task scales, and merging methods. As in common merging protocols, we chose these hyperparameters once on a validation subset of 256 images per task (ViT-B/32, 8 tasks) and kept them fixed for all other settings. Merge++ itself uses no real data. Full implementation details and the computational cost are given in Appendix~\ref{supp:impl}, and standard deviations over generation seeds in Appendix~\ref{supp:pertask}.
\subsection{Main Results}

Table~\ref{tab:main} reports the accuracy of each base merge and its improvement after applying Merge++; per-task accuracies and standard deviations over generation seeds are given in Appendix~\ref{supp:pertask}. The standard deviation of each gain over three generation seeds is at most 0.26 points. Merge++ improves the average accuracy of nearly every configuration, with average gains of $+3.9$, $+3.7$, and $+6.6$ points on ViT-B/32, $+2.2$, $+2.4$, and $+4.6$ on ViT-B/16, and $+1.8$, $+1.9$, and $+4.0$ on ViT-L/14 for 8, 14, and 20 tasks, respectively. In every architecture, the gains are largest at 20 tasks, consistent with interference between experts growing with their number. The gains decrease with model size, as larger merged models already retain more of the accuracy of their experts.

The improvement depends on the base merge. Task Arithmetic and TIES gain the most ($+3.2$ to $+7.6$), but spectral and optimisation-based merges also improve, with TSV-M gaining up to $+5.0$ and SWUDI-A up to $+3.0$, indicating that the signal recovered by Merge++ is complementary to what these methods achieve in weight space. The most pronounced case is WUDI at 20 tasks, whose accuracy collapses to 62.5, 69.8, and 82.6 on the three architectures, in line with the 61.1 reported for ViT-B/32 by \citet{wei2026swudi}. Merge++ restores it to 79.0, 81.2, and 89.2, on par with the strongest methods. Applied to the strongest base merge, Merge++ improves the best accuracy in 8 of the 9 settings, by up to $+3.0$ (ViT-B/32, 20 tasks).

The averages at 14 and 20 tasks are reduced by a single task, PCAM. On strong base merges, Merge++ lowers its accuracy, for instance by 15.2 points for SWUDI-A on ViT-B/16 with 14 tasks, the configuration that is otherwise unchanged; excluding PCAM, its average improves by $+1.2$. The synthetic PCAM images are classified correctly by the expert but at chance by the pretrained model, suggesting that these histopathology images lie too far from the natural-image domain for the expert's knowledge to transfer through inversion. We discuss this limitation in Sec.~\ref{sec:conclusion}.

\subsection{Qualitative Results}

Figure~\ref{fig:synth} shows synthetic images generated for the eight tasks with ViT-B/32. The images are not realistic reconstructions of the training data, and many are abstract. They nevertheless contain recognisable task-specific patterns, such as building and landscape fragments for SUN397, car parts for Cars, aerial structures for RESISC45, field patterns for EuroSAT, digits for SVHN and MNIST, traffic-sign shapes and symbols for GTSRB, and regular textures for DTD. This is consistent with the objective of the synthesis, which is not to reproduce the training images but to generate inputs that each expert classifies confidently and that activate the representations specific to its task.

\begin{table*}[t]
\centering
\caption{\textbf{Ablation of the synthesis terms.} Each row synthesises the images with a different subset of the terms in Eq.~\eqref{eq:total}, namely the confidence term $\mathcal{L}_{\mathrm{conf}}$, the subspace term $\mathcal{L}_{\mathrm{sub}}$, and the task identifiability term $\mathcal{L}_{\mathrm{id}}$. The resulting images are used to refine six base merges with the same distillation recipe. The first row reports the average accuracy (\%) of each base merge over the 8 tasks with ViT-B/32, and the remaining rows report the improvement after refinement. The best result for each base merge is shown in bold. All variants use a single generation seed and 600 generation steps (450 in Table~\ref{tab:main}).}
\label{tab:ablation_terms}
\setlength{\tabcolsep}{8pt}
\renewcommand{\arraystretch}{1.15}
\begin{tabular}{>{\columncolor{grayrow}}cc>{\columncolor{grayrow}}ccccccc}
\toprule
$\mathcal{L}_{\mathrm{conf}}$ & $\mathcal{L}_{\mathrm{sub}}$ & $\mathcal{L}_{\mathrm{id}}$ & TA & TIES & Iso-CTS & TSV-M & WUDI & SWUDI-A \\
\midrule
\multicolumn{3}{>{\columncolor{white}}c}{\textit{base merge}} & \textit{69.96} & \textit{73.33} & \textit{83.55} & \textit{83.71} & \textit{84.87} & \textit{85.56} \\
\midrule
\checkmark &  &  & +6.71 & +5.83 & +1.52 & +1.47 & +1.35 & +0.65 \\
\checkmark &  & \checkmark & +7.57 & +6.89 & +1.89 & +2.07 & +1.79 & +1.10 \\
\checkmark & \checkmark &  & +7.53 & +6.80 & +2.79 & +2.40 & +2.21 & +1.61 \\
\checkmark & \checkmark & \checkmark & \cellcolor{grayrow}\textbf{+8.00} & \cellcolor{grayrow}\textbf{+7.32} & \cellcolor{grayrow}\textbf{+2.85} & \cellcolor{grayrow}\textbf{+2.65} & \cellcolor{grayrow}\textbf{+2.29} & \cellcolor{grayrow}\textbf{+1.71} \\
\bottomrule
\end{tabular}
\end{table*}

\subsection{Analysis}

To understand where Merge++ acts, we compare the intermediate representations of each merged model with those of the experts. For every task, we pass 128 real test images through the merged model and through the expert of that task, and compute the cosine similarity between their class-token representations at the output of every transformer block, averaged over tasks. Real images are used here for analysis only. A similarity of 1 at a given block means that, up to that block, the merged model represents the images of each task exactly as its expert does; lower values indicate that merging has moved the representations of the merged model away from those of the experts. Figure~\ref{fig:analysis} shows how this similarity changes after applying Merge++, together with the relative change of the weights of each block. For readability, we show three of the six base merges, one arithmetic (Task Arithmetic), one spectral (TSV-M), and one optimisation-based (SWUDI-A); the remaining merges follow the same pattern and are shown in Appendix~\ref{supp:analysis}.

Before refinement, the similarity remains close to 1 in the early blocks, decreases slightly in the middle blocks, and drops sharply in the last few, reaching between 0.68 and 0.90 at the last block. The merged models thus deviate from their experts mostly in the final blocks, and more so for weaker merges. Merge++ raises the similarity mainly in these blocks and closes a substantial part of the remaining gap to the experts (17--43\% at the last block), for all architectures and base merges. The relative change of the weights (Figure~\ref{fig:analysis}, bottom) is below 1\% in every block, so refinement makes only small, targeted adjustments. The change grows towards the final blocks and is larger for merges that deviate more from their experts, so the weights change most where the representations deviate most.

\subsection{Ablation of the Synthesis Terms}

Our synthesis objective combines the subspace and task identifiability terms with the confidence term. To evaluate the contribution of each term, we synthesise images with each subset of the terms and refine all six base merges with each image set (Table~\ref{tab:ablation_terms}). Both terms improve over confidence-only synthesis on every base merge, and combining them yields the largest gain in all six cases. The two terms matter most where the base merge is already strong. On TSV-M, for example, confidence-only images improve the base merge by +1.47 points, the subspace term raises this to +2.40, and the full objective to +2.65; Iso-CTS, WUDI, and SWUDI-A follow the same pattern. On Task Arithmetic and TIES, confidence-only images already yield large gains (+6.71 and +5.83), and the two terms add a smaller, similar improvement. The subspace term is therefore the main source of improvement on the merges where confidence-only synthesis is least effective.

%% file: sec/5_conclusion.tex
\section{Conclusion}
\label{sec:conclusion}

Merging methods combine experts by operating on their weights, without seeing how the merged model behaves on inputs. Merge++ provides this signal without real data, by generating images from each expert and training the merged model to reproduce the experts on them. The difficulty is that images of different tasks update the same shared weights and can interfere with one another. Our analysis shows that refinement averages the experts wherever the images of several tasks overlap. Merge++ therefore generates images that use input directions of their own task and that the pretrained model recognises as belonging to that task. It consistently improves merges ranging from simple task arithmetic to recent spectral and optimisation-based methods. Its corrections are concentrated in the final blocks of the network, where the merged models deviate most from their experts. More broadly, our results show that data synthesised from the experts can complement weight-space merging, and that designing it to avoid interference between the updates of different tasks is central to its effectiveness.

\paragraph{Limitations and future work.} Image generation works less well for tasks far from natural images, such as histopathology, where the generated images do not resemble real data. Our image parameterisation and colour decorrelation encode statistics of natural images, and choosing priors suited to each task's domain may improve generation for such tasks. Image generation is also the main computational cost. It is incurred once per set of tasks and shared by all merges, yet faster generation would make the method cheaper to apply. Following the established setup of the merging literature, our evaluation considers CLIP vision encoders on image classification, and other architectures and tasks remain to be tested. Extending Merge++ to language models would require synthesis over discrete tokens, for instance on token embeddings with priors that keep them close to real tokens, which we leave for future work.

%% file: sec/9_llm_statement.tex
\section*{LLM Usage Statement}

We used GitHub Copilot CLI throughout this work. It implemented the method and experiments and replicated baselines, and, under the authors' direction, ran experiments. Under the authors' direction, it also proposed experiment variants and ablations and helped analyse their results. It further assisted with refining the authors' theoretical analysis, the literature search, drafting the paper, and \LaTeX{} and figure scripts. We also obtained reviewer-style feedback on the manuscript from Claude. The authors reviewed all AI-generated code and text, verified all citations and proofs, and take full responsibility for the content. All numbers were taken from evaluation script outputs and not fabricated or hallucinated by AI agents.

%% file: sec/A_appendix.tex
\section{Theoretical Analysis}\label{supp:theory}

We analyse refinement of a single linear layer with fixed inputs (Sec.~3.4) and make precise the two requirements: \textbf{(I)} the activations of task $t$ have energy along the directions on which $\tau_t$ acts, and \textbf{(II)} the activations of different tasks do not share directions on which their task vectors disagree. \Cref{supp:gd} gives the limit of gradient descent, \cref{supp:prop1} proves Proposition~1, and \cref{supp:general} removes the shared-eigenbasis assumption by stating (I) and (II) as conditions on the activation spans. \Cref{supp:direct} shows that the subspaces of Eq.~(3) are linearly independent and bounds the irreducible loss by the activation energy outside them. \Cref{supp:measure} evaluates this on the synthetic activations.

\subsection{Setting and Notation}\label{supp:setting}

A linear layer has weights $\theta_0$ (pretrained), $\theta_t=\theta_0+\tau_t$ (expert $t$) and $\theta_m$ (merge) in $\R^{p\times d}$. The rows of $A_t\in\R^{n_t\times d}$ are the input activations of the synthetic images of task $t$, one per token. Refinement minimises
\begin{equation}
    \mathcal{L}(\theta)=\sum_{t=1}^{T}\big\|A_t(\theta-\theta_t)^\top\big\|_F^2
    \label{eq:supp_loss}
\end{equation}
by gradient descent, $\theta^{(k+1)}=\theta^{(k)}-\eta\,\nabla\mathcal{L}(\theta^{(k)})$, from $\theta^{(0)}=\theta_m$. We write
\begin{equation}
    G_t=A_t^\top A_t,\quad G=\sum_{t}G_t,\quad B=\sum_{t}\theta_tG_t,\quad R_t=\range(G_t),\quad \Pi=GG^{+},
\end{equation}
where $G^{+}$ is the pseudoinverse of $G$. Thus $R_t$ is the span of the activations of task $t$, and $\Pi$ is the orthogonal projection onto $\range(G)=R_1+\dots+R_T$.

\subsection{Convergence of Gradient Descent}\label{supp:gd}

Both Proposition~1 and its extension rest on a closed form for the limit of gradient descent.

\begin{lemma}\label{lem:supp_gd}
For $0<\eta<1/\lambda_{\max}(G)$, gradient descent converges to the global minimiser
\begin{equation}
    \theta^\star=BG^{+}+\theta_m\,(I-\Pi).
    \label{eq:supp_limit}
\end{equation}
In particular, $\theta^\star c=\theta_m c$ for every $c\perp\range(G)$.
\end{lemma}

\begin{proof}
The gradient is $\nabla\mathcal{L}(\theta)=2(\theta G-B)$. Since $0\preceq G_t\preceq G$, each $G_t$ vanishes on $\range(G)^\perp$, so
\begin{equation*}
    B(I-\Pi)=0,\qquad \nabla\mathcal{L}(\theta)(I-\Pi)=0,\qquad \theta^{(k)}(I-\Pi)=\theta_m(I-\Pi)\ \ \forall k.
\end{equation*}
Since $G^{+}G=\Pi$, $B\Pi=B$ and $G^{+}(I-\Pi)=0$, we have $\theta^\star G=B$ and $\theta^\star(I-\Pi)=\theta_m(I-\Pi)$, and therefore
\begin{equation*}
    \theta^{(k+1)}-\theta^\star=\big(\theta^{(k)}-\theta^\star\big)(I-2\eta G).
\end{equation*}
The rows of $\theta^{(k)}-\theta^\star$ lie in $\range(G)$, where the eigenvalues of $I-2\eta G$ are $1-2\eta\lambda\in(-1,1)$. Hence $\theta^{(k)}\to\theta^\star$. As $\mathcal{L}$ is convex and $\nabla\mathcal{L}(\theta^\star)=0$, $\theta^\star$ is a global minimiser.
\end{proof}

\subsection{Proof of Proposition 1}\label{supp:prop1}

Proposition~1 assumes that the Gram matrices share an orthonormal eigenbasis $\{s_j\}_{j=1}^d$, \ie, $G_t=\sum_j g_{tj}\,s_js_j^\top$ with $g_{tj}=\|A_ts_j\|^2$ and $g_j=\sum_tg_{tj}$. The assumption decouples $\mathcal{L}$ into independent weighted least-squares problems, one per direction $s_j$, and so gives $\theta^\star$ and $\mathcal{L}(\theta^\star)$ explicitly in terms of the energies $g_{tj}$.

\begin{propone}
Gradient descent on~\eqref{eq:supp_loss}, with a step size as in \cref{lem:supp_gd}, converges to a minimiser $\theta^\star$ with
\begin{align}
    \theta^\star s_j&=\theta_m s_j && \text{if } g_j=0, \label{eq:supp_unchanged}\\[4pt]
    \theta^\star s_j&=\sum_{t=1}^{T}\frac{g_{tj}}{g_j}\,\theta_t s_j && \text{if } g_j>0, \label{eq:supp_average}
\end{align}
and the irreducible loss is
\begin{equation}
    \mathcal{L}(\theta^\star)=\sum_{j:\,g_j>0}\;\frac{1}{g_j}\sum_{t<u} g_{tj}\,g_{uj}\,\big\|(\tau_t-\tau_u)\,s_j\big\|^2 .
    \label{eq:supp_remaining}
\end{equation}
\end{propone}

\begin{proof}
In the basis $\{s_j\}$,
\begin{equation*}
    G=\sum_j g_j\,s_js_j^\top,\qquad G^{+}=\sum_{j:\,g_j>0}\frac{s_js_j^\top}{g_j},\qquad \Pi=\sum_{j:\,g_j>0}s_js_j^\top .
\end{equation*}
Applying~\eqref{eq:supp_limit} to $s_j$ gives
\begin{equation*}
    \theta^\star s_j=\begin{cases}\theta_m s_j, & g_j=0,\\[2pt] Bs_j/g_j=\sum_t \frac{g_{tj}}{g_j}\,\theta_ts_j, & g_j>0,\end{cases}
\end{equation*}
which are~\eqref{eq:supp_unchanged} and~\eqref{eq:supp_average}. The loss decouples as
\begin{equation*}
    \mathcal{L}(\theta)=\sum_j\sum_t g_{tj}\,\|\theta s_j-\theta_ts_j\|^2 .
\end{equation*}
Each inner sum is minimised at the weighted mean~\eqref{eq:supp_average}, where it equals
\begin{equation*}
    \sum_t g_{tj}\,\|\theta^\star s_j-\theta_ts_j\|^2=\frac{1}{g_j}\sum_{t<u}g_{tj}\,g_{uj}\,\|\theta_ts_j-\theta_us_j\|^2 .
\end{equation*}
Using $\theta_t-\theta_u=\tau_t-\tau_u$ and summing over $j$ gives~\eqref{eq:supp_remaining}.
\end{proof}

\noindent Requirements (I) and (II) follow from~\eqref{eq:supp_unchanged} and~\eqref{eq:supp_remaining}, which rely on the common eigenbasis. \Cref{supp:general} shows that both hold without it.

\subsection{Extension to Arbitrary Activations}\label{supp:general}

Without a common eigenbasis, $\mathcal{L}$ no longer splits over individual directions, so (I) and (II) cannot be read off a per-direction formula. Both requirements, however, concern only which directions the activations of each task occupy. They can therefore be stated through the spans $R_t$, without any assumption on the Gram matrices.

\begin{prop}\label{prop:supp_general}
Let the activations be arbitrary, and let $\Pi_t$ be the orthogonal projection onto $R_t$.
\begin{enumerate}
\item[(i)] If $\tau_t'\Pi_t=\tau_t\Pi_t$, replacing $\tau_t$ by $\tau_t'$ leaves $\theta^\star$ unchanged.
\item[(ii)] If $(\tau_t-\tau_u)\,r\neq0$ for some $r\in R_t\cap R_u$ with $t\neq u$, then $\mathcal{L}(\theta^\star)>0$.
\item[(iii)] If $R_1+\dots+R_T$ is a direct sum, then $\mathcal{L}(\theta^\star)=0$.
\end{enumerate}
\end{prop}

\begin{proof}
\emph{(i).} By~\eqref{eq:supp_limit}, $\theta^\star$ depends on the experts only through $B$. Since the columns of $G_t$ lie in $R_t$, $G_t=\Pi_tG_t$, so replacing $\tau_t$ by $\tau_t'$ changes $B$ by
\begin{equation*}
    (\tau_t'-\tau_t)\,G_t=(\tau_t'-\tau_t)\,\Pi_t\,G_t=0 .
\end{equation*}
For (ii) and (iii), note that $\mathcal{L}(\theta^\star)=\min_\theta\mathcal{L}(\theta)$ by \cref{lem:supp_gd}, and that
\begin{equation*}
    \mathcal{L}(\theta)=0\iff \theta r=\theta_t r\quad \forall\, r\in R_t,\ \forall\, t .
\end{equation*}
\noindent\emph{(ii).} Such a $\theta$ would give $\theta_t r=\theta r=\theta_u r$, \ie, $(\tau_t-\tau_u)\,r=0$, contradicting the assumption.

\noindent\emph{(iii).} Every $r\in R=\bigoplus_t R_t$ decomposes uniquely as $r=\sum_t r_t$ with $r_t\in R_t$. The linear map
\begin{equation*}
    \theta r=\sum_t\theta_t r_t\ \ (r\in R),\qquad \theta r=0\ \ (r\in R^\perp)
\end{equation*}
satisfies $\theta r=\theta_t r$ for all $r\in R_t$, so $\mathcal{L}(\theta)=0$.
\end{proof}

\noindent Part~(i) shows that refinement depends on expert $t$ only through its action on $R_t$, and by \cref{lem:supp_gd} it leaves the merge unchanged on $(R_1+\dots+R_T)^\perp$. This gives (I) for arbitrary activations, generalising~\eqref{eq:supp_unchanged}. By (ii), a shared direction on which two task vectors differ implies a positive irreducible loss, which gives (II) and generalises the pairwise terms of~\eqref{eq:supp_remaining}. By (iii), independent spans remove the irreducible loss entirely, and \cref{prop:supp_bound} makes this quantitative.

\subsection{Independence of the Task-Specific Subspaces}\label{supp:direct}

The subspace term concentrates the activations of each task in a subspace $S_t$. We show that the $S_t$ are linearly independent (\cref{lem:supp_direct}), so that Proposition~\ref{prop:supp_general}(iii) gives zero irreducible loss if the activations of each task lie in $S_t$. \Cref{prop:supp_bound} makes this quantitative, bounding the irreducible loss by the activation energy outside the $S_t$.

Eq.~(3) builds $S_t$ from the matrix $V_t\in\R^{d\times k}$ of the top-$k$ right singular vectors of $\tau_t$, by removing their components along the directions of the other tasks,
\begin{equation}
    S_t=\range\big((I-P_{W_t})V_t\big),
\end{equation}
where $W_t$ is the span of the columns of all $V_u$, $u\neq t$, and $P_{W_t}$ the orthogonal projection onto $W_t$. The $S_t$ are not orthogonal in general, but they are linearly independent.

\begin{lemma}\label{lem:supp_direct}
For any $V_1,\dots,V_T$, the sum $S_1+\dots+S_T$ is direct.
\end{lemma}

\begin{proof}
Let $\mathcal{V}$ be the span of the columns of all $V_t$. By construction, $S_u\subseteq\mathcal{V}$ and $S_u\perp W_u$. Let $z_t\in S_t$ with $\sum_t z_t=0$, fix $u$, and let $v$ lie in the span of $V_u$. Since $v\in W_t$ for every $t\neq u$,
\begin{equation*}
    0=v^\top\sum_t z_t=v^\top z_u .
\end{equation*}
Hence $z_u\perp V_u$, and also $z_u\perp V_t$ for $t\neq u$, so $z_u\perp\mathcal{V}$. Since $z_u\in\mathcal{V}$, $z_u=0$.
\end{proof}

\begin{corollary}\label{prop:supp_bound}
Let $P_t$ be the orthogonal projection onto $S_t$. There exists $\hat\theta$ with $\hat\theta z=\theta_tz$ for all $z\in S_t$ and all $t$, and
\begin{equation}
    \mathcal{L}(\theta^\star)\;\leq\;\sum_{t=1}^{T}\|\hat\theta-\theta_t\|_2^2\,\big\|A_t(I-P_t)\big\|_F^2 .
    \label{eq:supp_bound}
\end{equation}
\end{corollary}

\begin{proof}
$\hat\theta$ exists by \cref{lem:supp_direct}, as in the proof of Proposition~\ref{prop:supp_general}(iii). Since $\mathcal{L}(\theta^\star)\leq\mathcal{L}(\hat\theta)$ and $(\hat\theta-\theta_t)P_t=0$,
\begin{equation*}
    \big\|A_t(\hat\theta-\theta_t)^\top\big\|_F=\big\|A_t(I-P_t)(\hat\theta-\theta_t)^\top\big\|_F\leq\big\|A_t(I-P_t)\big\|_F\,\|\hat\theta-\theta_t\|_2 .
    \qedhere
\end{equation*}
\end{proof}

\noindent The irreducible loss is therefore controlled by the activation energy of each task outside its subspace, which is the quantity the subspace term targets.

\begin{table}[t]
\centering
\begin{minipage}[b]{0.47\linewidth}
\centering
\caption{Normalised irreducible loss $\mathcal{L}(\theta^\star)/\mathcal{L}(\theta_0)$, averaged over the 36 layers of ViT-B/32 (8 tasks). Lower is better.}
\label{tab:supp_irreducible}
\begin{tabular}{lc}
\toprule
Images & Irreducible loss \\
\midrule
Gaussian noise & 0.767 \\
$\mathcal{L}_{\mathrm{conf}}$ & 0.474 \\
$\mathcal{L}_{\mathrm{conf}}+\mathcal{L}_{\mathrm{id}}$ & 0.385 \\
$\mathcal{L}_{\mathrm{conf}}+\mathcal{L}_{\mathrm{sub}}$ & 0.253 \\
$\mathcal{L}_{\mathrm{conf}}+\mathcal{L}_{\mathrm{sub}}+\mathcal{L}_{\mathrm{id}}$ & 0.246 \\
\midrule
Real validation images & 0.230 \\
\bottomrule
\end{tabular}
\end{minipage}\hfill
\begin{minipage}[b]{0.49\linewidth}
\centering
\caption{Fraction of the mean-centred activation energy of each task in its own subspace, $\|A_tP_t\|_F^2/\|A_t\|_F^2$, and in the subspaces of the other tasks, $\|A_tP_u\|_F^2/\|A_t\|_F^2$ averaged over $u\neq t$, both averaged over layers and tasks (ViT-B/32, 8 tasks). Ratio: own over other.}
\label{tab:supp_share}
\begin{tabular}{lccc}
\toprule
Images & Own $S_t$ & Other $S_u$ & Ratio \\
\midrule
Gaussian noise & 0.021 & 0.021 & 1.00 \\
$\mathcal{L}_{\mathrm{conf}}$ & 0.029 & 0.023 & 1.27 \\
$\mathcal{L}_{\mathrm{conf}}+\mathcal{L}_{\mathrm{id}}$ & 0.032 & 0.022 & 1.43 \\
$\mathcal{L}_{\mathrm{conf}}+\mathcal{L}_{\mathrm{sub}}$ & 0.067 & 0.021 & 3.25 \\
$\mathcal{L}_{\mathrm{conf}}+\mathcal{L}_{\mathrm{sub}}+\mathcal{L}_{\mathrm{id}}$ & 0.067 & 0.021 & 3.27 \\
\midrule
Real validation images & 0.052 & 0.021 & 2.47 \\
\bottomrule
\end{tabular}
\end{minipage}

\vspace{1.2em}
\includegraphics[width=\linewidth]{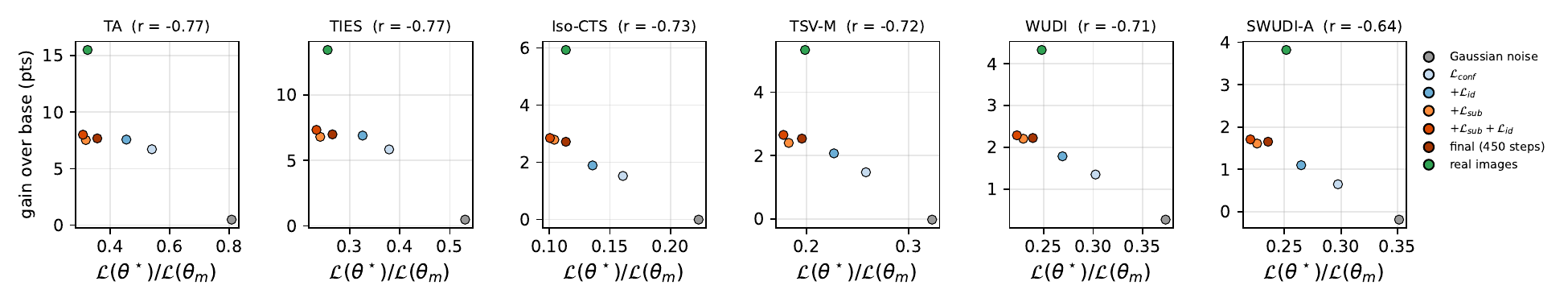}
\captionof{figure}{Accuracy gain of refinement against the irreducible loss normalised by the loss of the base merge, $\mathcal{L}(\theta^\star)/\mathcal{L}(\theta_m)$, for each base merge and image set (ViT-B/32, 8 tasks).}
\label{fig:supp_irreducible}
\end{table}

\subsection{Irreducible Loss of the Synthetic Activations}\label{supp:measure}

\cref{prop:supp_bound} predicts that concentrating the activations of each task in $S_t$ lowers the irreducible loss. We test this on the synthetic activations. As $\mathcal{L}$ depends on the weights only through $\theta-\theta_t$, we set $\theta_0=0$ without loss of generality, so that $\theta_t=\tau_t$ and, by~\eqref{eq:supp_limit},
\begin{equation}
    \mathcal{L}(\theta^\star)=\sum_{t=1}^{T}\tr\big(\tau_tG_t\tau_t^\top\big)-\tr\big(BG^{+}B^\top\big).
    \label{eq:supp_general_loss}
\end{equation}
We normalise by $\mathcal{L}(\theta_0)=\sum_t\tr(\tau_tG_t\tau_t^\top)$, the loss of the pretrained weights on the synthetic activations. The ratio $\mathcal{L}(\theta^\star)/\mathcal{L}(\theta_0)$ is the fraction of this loss that remains at the optimum. We compute it on the activations of $\theta_0$ for the query--key--value and two MLP layers of the 12 blocks of ViT-B/32 (8 tasks) and average over layers.

The subspace term lowers the normalised irreducible loss of confidence-only synthesis from 0.47 to 0.25, close to the 0.23 of real images, whereas the identifiability term alone lowers it to 0.39 (\cref{tab:supp_irreducible}). It raises the fraction of energy in each task's own subspace from 0.029 to 0.067, while the fraction in the subspaces of the other tasks stays at about 0.021, the level of Gaussian noise (\cref{tab:supp_share}). To relate the irreducible loss to the accuracy gain, \cref{fig:supp_irreducible} normalises it by the loss of the base merge, $\mathcal{L}(\theta_m)$. For every base merge, synthetic image sets with a lower normalised irreducible loss give a larger gain, consistent with (II).

\section{Implementation Details}\label{supp:impl}
\subsection{Experts and Benchmarks}\label{supp:experts}

\paragraph{Models.} The pretrained models are the CLIP image encoders ViT-B/32, ViT-B/16 and ViT-L/14 with the original OpenAI weights~\cite{radford2021clip}, loaded with \texttt{open\_clip} and the QuickGELU activation used during their training. The experts are the fine-tuned encoders released by \citet{wang2024tall} for all twenty tasks and all three architectures. All methods in our comparison use these checkpoints.

\paragraph{Classification heads.} Each task has a fixed linear head whose rows are the CLIP text embeddings of its class names, averaged over the prompt templates of the task~\cite{radford2021clip, ilharco2023task}. The heads are released together with the checkpoints, are not changed during fine-tuning, and are shared by the pretrained, fine-tuned, merged and refined encoders. As in prior work on model merging~\cite{ilharco2023task, yadav2023ties, wang2024tall, tang2025tsvm, cheng2025wudi}, we assume they are available.

\paragraph{Benchmarks.} We use the three task sets of \citet{wang2024tall}, listed in \cref{tab:supp_tasks}. Accuracy is the top-1 accuracy on the test split of each task, averaged over the tasks of the benchmark. Following \citet{ilharco2023task}, a validation split is held out from the training set of each task, with $10\%$ of its images and at most 5{,}000. The validation split is used only for the coefficient selection of the baselines and for the selection of our hyperparameters, and never during inversion or distillation.

\begin{table}[h]
\centering
\caption{Tasks of the three benchmarks. Each benchmark contains the tasks of the smaller ones.}
\label{tab:supp_tasks}
\begin{tabular}{ll}
\toprule
Benchmark & Tasks \\
\midrule
8 tasks & SUN397, Cars, RESISC45, EuroSAT, SVHN, GTSRB, MNIST, DTD \\
14 tasks & + Flowers102, PCAM, FER2013, OxfordPets, STL10, CIFAR100 \\
20 tasks & + CIFAR10, Food101, FashionMNIST, EMNIST, KMNIST, RenderedSST2 \\
\bottomrule
\end{tabular}
\end{table}

\subsection{Baseline Protocols}\label{supp:baselines}

We run all six baselines on the same checkpoints, using the official implementations of their authors or faithful ports of them, with the hyperparameters of the original papers. For TIES, this means keeping the top 20\% of the task-vector magnitudes, as reported by the authors. Every merge has the form $\theta_m=\theta_0+\alpha\,\tau_m$, where $\tau_m$ is the merged task vector and $\alpha$ a scaling coefficient. For Task Arithmetic, TIES and Iso-CTS, $\alpha$ is selected on the validation split. We evaluate a grid of 30 values with the selection loop of the official code and keep the one with the highest average normalised validation accuracy, \ie, the accuracy of the merge divided by that of the expert, averaged over tasks. TSV-M, WUDI and SWUDI-A use $\alpha=1$, as reported by the authors.

Merge++ is applied to each resulting merge as it is, without changing its coefficient or any other setting.

\subsection{Inversion}\label{supp:inversion}

\paragraph{Target classes.} Each task receives $M=256$ images, except SUN397, which receives 512 so that each of its 397 classes is targeted at least once. The target classes $y$ are assigned to the images in turn, cycling through the classes of the task.

\paragraph{Image parameterisation.} Each image is represented by a Laplacian pyramid~\cite{cazenavette2025lgm} with levels at resolutions $1,2,4,\dots,128$ and $224$. The optimisation starts with the coarsest level only, and a new level is added every 50 steps, so that the full pyramid is reached after 400 steps. When a level is added, the existing levels are rescaled so that the decoded image is unchanged, and the optimiser state is reset. The decoded image is mapped to colours with the decorrelation matrix of \citet{olah2017feature} and passed through $\mathrm{sigmoid}(2x)$, as in the implementation of \citet{cazenavette2025lgm}.

\paragraph{Augmentation.} At every step, each image is augmented into $K=10$ copies. Each copy is independently flipped horizontally with probability $0.5$, cropped to a random region covering between $50\%$ and $100\%$ of its area, with aspect ratio between $3/4$ and $4/3$, and resized to $224\times224$. Gaussian noise is then added with probability $0.5$.

\paragraph{Objective.} The confidence term uses all $K$ copies. To reduce memory, the subspace and task identifiability terms are computed on two of the copies, which share a single forward pass of $\theta_0$. The set $\mathcal{L}$ contains the query--key--value projection and the two MLP layers of every transformer block, \ie, 36 layers for ViT-B/32 and ViT-B/16, and 72 for ViT-L/14. In the subspace term, the activations are mean-centred over all tokens of the batch. The adaptive weights of Eq.~(7) are recomputed at every step, with $\lambda=0.3$ and $k=16$. Results for other values of $k$ are given in \cref{supp:kablation}.

\paragraph{Optimisation.} The images are optimised with Adam at a constant learning rate of $2\times10^{-3}$ for 450 steps, in full precision. They are processed in batches of 64, 16 and 4 images for ViT-B/32, ViT-B/16 and ViT-L/14, respectively, and the gradient norms of Eq.~(7) are computed per batch. The main results are averaged over three generation seeds.

\subsection{Distillation}\label{supp:distill}

\paragraph{Trainable parameters.} The student is initialised from the merge and only the weight matrices of the four linear layers in each transformer block are trained, \ie, the query--key--value and output projections of the attention and the two MLP layers. All other parameters, including biases, normalisation layers and embeddings, remain those of the merge. The experts and the classification heads are frozen.

\paragraph{Data.} Each batch is sampled uniformly at random from the union of the synthetic sets of all tasks, and each image is paired with the expert that generated it. Eight images per task are held out to monitor training and are not used as training data. Each image is augmented with a random crop covering between $60\%$ and $100\%$ of its area, resized to $224\times224$, a random horizontal flip, and colour jitter of brightness, contrast and saturation with strength $0.2$.

\paragraph{Optimisation.} We train for 600 steps with AdamW, a learning rate of $2\times10^{-6}$ with cosine decay to zero, weight decay $0.01$, and gradient clipping at norm $1.0$. The batch size is 1024 for all architectures, and we use gradient accumulation where memory requires it. The temperature of the logit matching loss is set to $\kappa=2$.

\subsection{Computational Cost}\label{supp:compute}

All experiments were conducted on a system with four RTX 6000 Ada GPUs. \Cref{tab:supp_compute} reports the GPU-hours of both stages. Image generation is performed once for a set of tasks, and the same images serve every base merge that is refined afterwards. The refined model has the same architecture and size as the base merge, so Merge++ adds no cost at inference.

\begin{table}[H]
\centering
\caption{Average computational cost of Merge++ in GPU-hours on an RTX 6000 Ada GPU, for generating the images of one task and for refining one base merge.}
\label{tab:supp_compute}
\begin{tabular}{lcc}
\toprule
Model & Generation & Distillation \\
\midrule
ViT-B/32 & 0.65 & 0.7 \\
ViT-B/16 & 2.3 & 1.8 \\
ViT-L/14 & 10.3 & 6.2 \\
\bottomrule
\end{tabular}
\end{table}

\section{Per-Task Results}\label{supp:pertask}

Tables~\ref{tab:supp_pertask_vitb32} to~\ref{tab:supp_pertask_vitl14} report the accuracy on every task for each architecture, task scale and base merge, before and after refinement with Merge++.

Merge++ closes a substantial part of the gap to the experts on most tasks. The main exception is PCAM. There, refinement moves the accuracy of every base merge towards a similar level, raising weak merges and lowering strong ones. This suggests that the synthetic images of PCAM do not capture what separates its classes in real data. Its images are small histology patches that differ from CLIP's pretraining data, and their classes differ in fine local detail, which the image prior of Merge++, designed for natural images, does not reproduce well. On SVHN and FashionMNIST, refinement changes the accuracy only slightly.

\input{figs/supp/pertask_ViT-B-32}
\input{figs/supp/pertask_ViT-B-16}
\input{figs/supp/pertask_ViT-L-14}
\clearpage

\section{Analysis for the Remaining Base Merges}\label{supp:analysis}

Figure~\ref{fig:supp_analysis} repeats the analysis of Section~4.3 for the three base merges not shown in Figure~3, TIES, Iso-CTS and WUDI, with the same protocol. These merges follow the same pattern as the other three. Before refinement, the similarity to the experts remains close to 1 in the early blocks and drops mainly in the last few, reaching between 0.70 and 0.89 at the last block. Merge++ raises the similarity mainly in these blocks and closes 22--43\% of the remaining gap at the last block. The relative change of the weights is below 0.9\% in every block and is largest in the final blocks.

\input{figs/supp/fig_analysis_remaining}

\section{Additional Ablations}\label{supp:ablations}
\subsection{Subspace Dimension}\label{supp:kablation}

We vary the dimension $k$ of the task-specific subspaces $S_t^\ell$ (Eq.~(3)) with ViT-B/32 on 8 tasks, keeping all other settings fixed. With 8 tasks, $Tk$ does not exceed the input dimension of any layer for $k$ up to 64. Table~\ref{tab:supp_k} reports the gain of Merge++ over each base merge. The gain increases from $k=4$ to $k=16$ for every base merge and changes little for larger $k$. The value $k=16$, used in all other experiments, gives the largest gain for five of the six base merges.

\begin{table}[H]
\centering
\caption{Gain in average top-1 accuracy (\%) of Merge++ over each base merge for different subspace dimensions $k$, with ViT-B/32 on 8 tasks. The first row gives the accuracy of the base merges.}
\label{tab:supp_k}
\setlength{\tabcolsep}{5pt}
\begin{tabular}{lcccccc}
\toprule
& TA & TIES & Iso-CTS & TSV-M & WUDI & SWUDI-A \\
\midrule
base & 70.0 & 73.3 & 83.5 & 83.7 & 84.9 & 85.6 \\
\midrule
$k=4$ & +6.8 & +6.4 & +2.2 & +2.1 & +2.0 & +1.5 \\
$k=8$ & +7.0 & +6.6 & +2.5 & +2.3 & +2.1 & +1.5 \\
\rowcolor{grayrow}
$k=16$ & \textbf{+7.6} & +7.0 & \textbf{+2.6} & \textbf{+2.5} & \textbf{+2.2} & \textbf{+1.6} \\
$k=32$ & +7.4 & +6.7 & +2.5 & +2.2 & +2.1 & \textbf{+1.6} \\
$k=64$ & \textbf{+7.6} & \textbf{+7.1} & +2.5 & +2.3 & +2.0 & +1.5 \\
\bottomrule
\end{tabular}
\end{table}

\subsection{Inputs to Distillation}

To measure how much of the gain comes from the synthetic images, we replace them by other inputs while keeping the distillation unchanged, with ViT-B/32 on 8 tasks. We consider Gaussian noise, with pixels drawn independently from $\mathcal{N}(0.5, 0.25^2)$ and clipped to $[0,1]$, and unlabelled real images from the validation split of each task. Each set has the same number of images per task as the synthetic set, and its images are distilled from the expert of their task. Table~\ref{tab:supp_inputs} reports the results. Real images give the largest gains, and the synthetic images recover 42--52\% of them without any real data.

\begin{table}[H]
\centering
\caption{Gain in average top-1 accuracy (\%) of Merge++ over each base merge when distilling on different inputs, with ViT-B/32 on 8 tasks. The first row gives the accuracy of the base merges. Real images are taken from the validation split and used without labels.}
\label{tab:supp_inputs}
\setlength{\tabcolsep}{5pt}
\begin{tabular}{lcccccc}
\toprule
& TA & TIES & Iso-CTS & TSV-M & WUDI & SWUDI-A \\
\midrule
base & 70.0 & 73.3 & 83.5 & 83.7 & 84.9 & 85.6 \\
\midrule
Gaussian noise & +0.5 & +0.5 & 0.0 & 0.0 & +0.3 & --0.2 \\
\rowcolor{grayrow}
Synthetic (Merge++) & +7.6 & +7.0 & +2.6 & +2.5 & +2.2 & +1.6 \\
\midrule
Real (unlabelled) & +15.5 & +13.4 & +5.9 & +5.3 & +4.3 & +3.8 \\
\bottomrule
\end{tabular}
\end{table}

\subsection{Initialisation from the Zero-Shot Model}

Merge++ transfers knowledge from the experts in two ways. The base merge carries their weights and the synthetic images carry what distillation adds. To isolate the contribution of the images, we apply Merge++ to the zero-shot CLIP model $\theta_0$ instead of a base merge, with the same synthetic images and settings on 8 tasks. Table~\ref{tab:supp_zeroshot} reports the results. Merge++ improves the zero-shot model by 11.6--14.8 points, which shows that the synthetic images alone transfer substantial knowledge from the experts. As expected, the result stays below the refined base merges because the zero-shot model starts without the weights of the experts. The two sources are therefore complementary, and Merge++ is most effective when applied to a merge.

\begin{table}[H]
\centering
\caption{Average top-1 accuracy (\%) of the zero-shot model and its improvement after applying Merge++, on 8 tasks.}
\label{tab:supp_zeroshot}
\setlength{\tabcolsep}{5pt}
\begin{tabular}{llc>{\hspace{-17pt}}lc>{\hspace{-17pt}}lc>{\hspace{-17pt}}l}
\toprule
& & ViT-B/32 & & ViT-B/16 & & ViT-L/14 & \\
\midrule
Zero-shot & base & 48.0 & & 55.4 & & 65.3 & \\
\rowcolor{grayrow}
& {\footnotesize\textbf{w/ Merge++}} & 62.8 & {\footnotesize\textcolor{posgreen}{\textbf{+14.8}}} & 68.3 & {\footnotesize\textcolor{posgreen}{\textbf{+12.8}}} & 76.9 & {\footnotesize\textcolor{posgreen}{\textbf{+11.6}}} \\
\bottomrule
\end{tabular}
\end{table}

\section{Additional Qualitative Results}\label{supp:qual}

\begin{figure}[p]
\centering
\includegraphics[width=\linewidth]{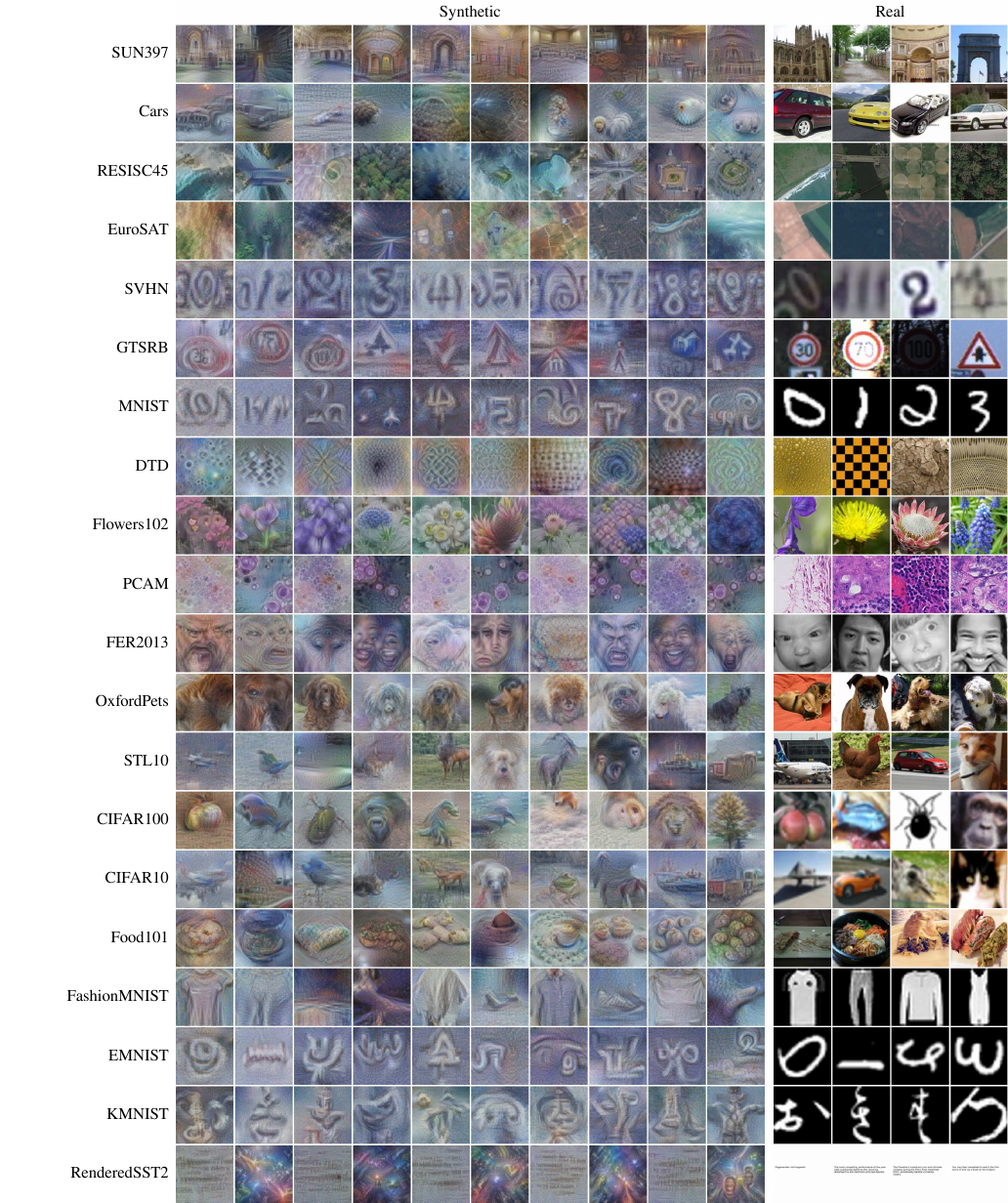}
\caption{Synthetic images generated by Merge++ from the ViT-B/32 experts of the 20-task benchmark, without any real data (left), and real validation images of the same tasks for reference (right). Each row shows one task. For tasks with fewer classes than columns, classes are repeated with different images.}
\label{fig:supp_synth_a}
\end{figure}

Figure~\ref{fig:supp_synth_a} shows synthetic images generated by Merge++ from the ViT-B/32 experts of the 20-task benchmark, together with real images from the validation split of each task.

\clearpage

%% file: figs/supp/pertask_ViT-B-32.tex
\begin{table}[p]
\centering
\caption{Per-task top-1 accuracy (\%) with ViT-B/32 of each base merge and its improvement after applying Merge++, reported as the mean $\pm$ standard deviation over generation seeds. Expert denotes the fine-tuned expert of each task.}
\label{tab:supp_pertask_vitb32}
\scriptsize\setlength{\tabcolsep}{3pt}
\resizebox{\linewidth}{!}{%
\begin{tabular}{l|c|cc|cc|cc|cc|cc|cc}
\toprule
 & & \multicolumn{2}{c|}{\textbf{TA} {\tiny\color{gray}[ICLR'23]}} & \multicolumn{2}{c|}{\textbf{TIES} {\tiny\color{gray}[NeurIPS'23]}} & \multicolumn{2}{c|}{\textbf{Iso-CTS} {\tiny\color{gray}[ICML'25]}} & \multicolumn{2}{c|}{\textbf{TSV-M} {\tiny\color{gray}[CVPR'25]}} & \multicolumn{2}{c|}{\textbf{WUDI} {\tiny\color{gray}[ICML'25]}} & \multicolumn{2}{c}{\textbf{SWUDI-A} {\tiny\color{gray}[arXiv'26]}} \\
\textbf{Task} & Expert & base & {\tiny\textbf{w/ Merge++}} & base & {\tiny\textbf{w/ Merge++}} & base & {\tiny\textbf{w/ Merge++}} & base & {\tiny\textbf{w/ Merge++}} & base & {\tiny\textbf{w/ Merge++}} & base & {\tiny\textbf{w/ Merge++}} \\
\midrule[\heavyrulewidth]
\multicolumn{14}{l}{\textbf{8 tasks}} \\
\midrule
SUN397 & 75.3 & 61.8 & \spos{7.0}{\tiny$\pm$0.07} & 62.6 & \spos{8.1}{\tiny$\pm$0.08} & 69.2 & \spos{3.5}{\tiny$\pm$0.07} & 67.2 & \spos{4.7}{\tiny$\pm$0.10} & 68.5 & \spos{4.0}{\tiny$\pm$0.08} & 69.7 & \spos{3.2}{\tiny$\pm$0.07} \\
Cars & 77.9 & 61.1 & \spos{3.3}{\tiny$\pm$0.59} & 62.7 & \spos{4.7}{\tiny$\pm$0.54} & 71.3 & \spos{4.1}{\tiny$\pm$0.45} & 70.6 & \spos{4.3}{\tiny$\pm$0.24} & 71.8 & \spos{3.0}{\tiny$\pm$0.24} & 71.0 & \spos{3.3}{\tiny$\pm$0.12} \\
RESISC45 & 95.9 & 71.5 & \spos{10.5}{\tiny$\pm$0.37} & 74.4 & \spos{11.1}{\tiny$\pm$0.33} & 88.5 & \spos{4.1}{\tiny$\pm$0.09} & 86.1 & \spos{5.2}{\tiny$\pm$0.18} & 85.5 & \spos{5.8}{\tiny$\pm$0.18} & 88.2 & \spos{3.9}{\tiny$\pm$0.06} \\
EuroSAT & 99.9 & 75.9 & \spos{12.8}{\tiny$\pm$0.38} & 79.0 & \spos{12.4}{\tiny$\pm$0.23} & 93.6 & \spos{3.1}{\tiny$\pm$0.30} & 94.4 & \spos{1.6}{\tiny$\pm$0.22} & 95.5 & \spos{2.1}{\tiny$\pm$0.04} & 96.6 & \spos{0.7}{\tiny$\pm$0.14} \\
SVHN & 97.6 & 74.6 & \spos{0.3}{\tiny$\pm$0.20} & 83.0 & \sneg{--3.7}{\tiny$\pm$0.28} & 84.7 & \sneg{--2.8}{\tiny$\pm$0.21} & 91.0 & \sneg{--4.1}{\tiny$\pm$0.10} & 93.8 & \sneg{--3.5}{\tiny$\pm$0.28} & 93.5 & \sneg{--3.4}{\tiny$\pm$0.12} \\
GTSRB & 99.1 & 67.4 & \spos{14.7}{\tiny$\pm$1.16} & 71.1 & \spos{12.6}{\tiny$\pm$0.99} & 91.5 & \spos{3.8}{\tiny$\pm$0.40} & 92.3 & \spos{2.9}{\tiny$\pm$0.43} & 94.5 & \spos{2.2}{\tiny$\pm$0.26} & 94.0 & \spos{2.4}{\tiny$\pm$0.28} \\
MNIST & 99.7 & 95.2 & \spos{2.1}{\tiny$\pm$0.22} & 98.0 & \spos{0.4}{\tiny$\pm$0.11} & 98.9 & \spos{0.2}{\tiny$\pm$0.03} & 99.3 & \szero{0.0}{\tiny$\pm$0.06} & 99.4 & \szero{0.0}{\tiny$\pm$0.06} & 99.5 & \szero{0.0}{\tiny$\pm$0.01} \\
DTD & 79.0 & 52.2 & \spos{10.0}{\tiny$\pm$0.24} & 55.9 & \spos{10.4}{\tiny$\pm$0.14} & 70.7 & \spos{4.8}{\tiny$\pm$0.35} & 68.7 & \spos{5.4}{\tiny$\pm$0.35} & 69.9 & \spos{4.1}{\tiny$\pm$0.08} & 72.1 & \spos{2.7}{\tiny$\pm$0.20} \\
\midrule
\rowcolor{grayrow}\textbf{Avg.} & 90.6 & 70.0 & \spos{7.6}{\tiny$\pm$0.26} & 73.3 & \spos{7.0}{\tiny$\pm$0.16} & 83.5 & \spos{2.6}{\tiny$\pm$0.16} & 83.7 & \spos{2.5}{\tiny$\pm$0.06} & 84.9 & \spos{2.2}{\tiny$\pm$0.02} & 85.6 & \spos{1.6}{\tiny$\pm$0.06} \\
\midrule[\heavyrulewidth]
\multicolumn{14}{l}{\textbf{14 tasks}} \\
\midrule
SUN397 & 75.3 & 63.9 & \spos{2.9}{\tiny$\pm$0.12} & 64.7 & \spos{3.3}{\tiny$\pm$0.07} & 68.8 & \spos{2.8}{\tiny$\pm$0.05} & 65.0 & \spos{4.7}{\tiny$\pm$0.07} & 62.0 & \spos{8.0}{\tiny$\pm$0.03} & 68.3 & \spos{2.8}{\tiny$\pm$0.03} \\
Cars & 77.9 & 59.5 & \sneg{--0.6}{\tiny$\pm$1.14} & 60.0 & \spos{1.3}{\tiny$\pm$1.01} & 68.1 & \spos{4.9}{\tiny$\pm$0.19} & 60.0 & \spos{10.5}{\tiny$\pm$0.58} & 61.6 & \spos{9.2}{\tiny$\pm$0.46} & 66.1 & \spos{5.4}{\tiny$\pm$0.41} \\
RESISC45 & 95.9 & 67.5 & \spos{6.2}{\tiny$\pm$0.16} & 69.8 & \spos{7.7}{\tiny$\pm$0.07} & 85.5 & \spos{4.0}{\tiny$\pm$0.04} & 80.1 & \spos{7.9}{\tiny$\pm$0.22} & 73.0 & \spos{14.1}{\tiny$\pm$0.13} & 83.2 & \spos{5.6}{\tiny$\pm$0.30} \\
EuroSAT & 99.9 & 62.5 & \spos{12.2}{\tiny$\pm$0.49} & 66.8 & \spos{13.7}{\tiny$\pm$0.41} & 90.2 & \spos{2.0}{\tiny$\pm$0.42} & 91.1 & \spos{1.0}{\tiny$\pm$0.07} & 87.2 & \spos{6.1}{\tiny$\pm$0.15} & 92.3 & \spos{1.5}{\tiny$\pm$0.09} \\
SVHN & 97.6 & 52.1 & \spos{9.4}{\tiny$\pm$0.12} & 57.5 & \spos{7.1}{\tiny$\pm$0.31} & 78.0 & \sneg{--0.6}{\tiny$\pm$0.25} & 83.9 & \sneg{--2.2}{\tiny$\pm$0.22} & 85.9 & \sneg{--0.9}{\tiny$\pm$0.17} & 87.7 & \sneg{--3.1}{\tiny$\pm$0.15} \\
GTSRB & 99.1 & 47.2 & \spos{18.0}{\tiny$\pm$0.78} & 55.1 & \spos{15.0}{\tiny$\pm$0.70} & 83.6 & \spos{6.7}{\tiny$\pm$0.54} & 84.9 & \spos{6.2}{\tiny$\pm$0.48} & 85.7 & \spos{6.4}{\tiny$\pm$0.44} & 88.2 & \spos{4.1}{\tiny$\pm$0.35} \\
MNIST & 99.7 & 80.1 & \spos{11.6}{\tiny$\pm$0.08} & 88.1 & \spos{6.7}{\tiny$\pm$0.09} & 97.9 & \spos{0.7}{\tiny$\pm$0.07} & 99.0 & \spos{0.2}{\tiny$\pm$0.05} & 99.0 & \spos{0.1}{\tiny$\pm$0.06} & 99.4 & \szero{0.0}{\tiny$\pm$0.06} \\
DTD & 79.0 & 48.3 & \spos{6.4}{\tiny$\pm$0.45} & 51.1 & \spos{6.6}{\tiny$\pm$0.27} & 66.4 & \spos{5.0}{\tiny$\pm$0.45} & 64.6 & \spos{5.0}{\tiny$\pm$0.43} & 60.9 & \spos{7.7}{\tiny$\pm$0.16} & 66.4 & \spos{3.3}{\tiny$\pm$0.24} \\
Flowers102 & 90.4 & 67.5 & \spos{1.7}{\tiny$\pm$0.45} & 64.8 & \spos{4.2}{\tiny$\pm$0.46} & 80.6 & \spos{3.6}{\tiny$\pm$0.08} & 74.6 & \spos{7.0}{\tiny$\pm$0.04} & 73.1 & \spos{8.1}{\tiny$\pm$0.16} & 79.2 & \spos{2.9}{\tiny$\pm$0.05} \\
PCAM & 87.9 & 64.0 & \spos{5.9}{\tiny$\pm$1.36} & 68.4 & \sneg{--0.9}{\tiny$\pm$2.43} & 82.2 & \sneg{--7.9}{\tiny$\pm$1.03} & 84.3 & \sneg{--10.9}{\tiny$\pm$0.68} & 84.0 & \sneg{--9.3}{\tiny$\pm$1.00} & 81.3 & \sneg{--10.7}{\tiny$\pm$0.86} \\
FER2013 & 72.9 & 42.4 & \spos{4.4}{\tiny$\pm$0.55} & 45.0 & \spos{4.7}{\tiny$\pm$0.53} & 57.5 & \spos{6.5}{\tiny$\pm$0.28} & 63.3 & \spos{3.0}{\tiny$\pm$0.24} & 63.1 & \spos{3.5}{\tiny$\pm$0.16} & 66.7 & \spos{1.0}{\tiny$\pm$0.20} \\
OxfordPets & 91.3 & 87.5 & \spos{0.4}{\tiny$\pm$0.48} & 86.8 & \spos{1.1}{\tiny$\pm$0.38} & 90.2 & \spos{0.6}{\tiny$\pm$0.11} & 89.9 & \spos{0.9}{\tiny$\pm$0.09} & 88.1 & \spos{2.0}{\tiny$\pm$0.06} & 90.3 & \spos{0.3}{\tiny$\pm$0.22} \\
STL10 & 98.0 & 96.6 & \szero{0.0}{\tiny$\pm$0.09} & 95.8 & \spos{0.7}{\tiny$\pm$0.06} & 97.8 & \sneg{--0.2}{\tiny$\pm$0.08} & 97.0 & \spos{0.5}{\tiny$\pm$0.09} & 95.9 & \spos{1.6}{\tiny$\pm$0.13} & 97.4 & \szero{0.0}{\tiny$\pm$0.06} \\
CIFAR100 & 89.3 & 69.5 & \spos{2.3}{\tiny$\pm$0.18} & 69.4 & \spos{3.5}{\tiny$\pm$0.29} & 76.3 & \spos{3.0}{\tiny$\pm$0.17} & 71.1 & \spos{5.2}{\tiny$\pm$0.27} & 68.3 & \spos{9.0}{\tiny$\pm$0.33} & 71.5 & \spos{5.1}{\tiny$\pm$0.46} \\
\midrule
\rowcolor{grayrow}\textbf{Avg.} & 89.6 & 64.9 & \spos{5.8}{\tiny$\pm$0.08} & 67.4 & \spos{5.3}{\tiny$\pm$0.19} & 80.2 & \spos{2.2}{\tiny$\pm$0.04} & 79.2 & \spos{2.8}{\tiny$\pm$0.11} & 77.7 & \spos{4.7}{\tiny$\pm$0.12} & 81.3 & \spos{1.3}{\tiny$\pm$0.11} \\
\midrule[\heavyrulewidth]
\multicolumn{14}{l}{\textbf{20 tasks}} \\
\midrule
SUN397 & 75.3 & 63.8 & \spos{2.3}{\tiny$\pm$0.12} & 64.1 & \spos{3.0}{\tiny$\pm$0.10} & 64.6 & \spos{5.0}{\tiny$\pm$0.11} & 62.6 & \spos{6.0}{\tiny$\pm$0.03} & 44.2 & \spos{22.5}{\tiny$\pm$0.16} & 66.6 & \spos{3.4}{\tiny$\pm$0.09} \\
Cars & 77.9 & 58.1 & \sneg{--0.7}{\tiny$\pm$2.08} & 57.1 & \spos{1.5}{\tiny$\pm$2.02} & 56.1 & \spos{10.7}{\tiny$\pm$0.93} & 49.3 & \spos{16.1}{\tiny$\pm$1.48} & 34.3 & \spos{26.2}{\tiny$\pm$1.57} & 56.1 & \spos{11.5}{\tiny$\pm$0.91} \\
RESISC45 & 95.9 & 64.5 & \spos{6.6}{\tiny$\pm$0.06} & 66.1 & \spos{7.8}{\tiny$\pm$0.19} & 78.7 & \spos{10.0}{\tiny$\pm$0.09} & 73.8 & \spos{12.8}{\tiny$\pm$0.23} & 50.7 & \spos{32.7}{\tiny$\pm$0.19} & 78.8 & \spos{8.6}{\tiny$\pm$0.24} \\
EuroSAT & 99.9 & 56.7 & \spos{12.8}{\tiny$\pm$1.04} & 58.8 & \spos{15.7}{\tiny$\pm$1.22} & 82.1 & \spos{8.3}{\tiny$\pm$0.48} & 84.6 & \spos{7.0}{\tiny$\pm$0.51} & 62.6 & \spos{26.7}{\tiny$\pm$0.62} & 88.3 & \spos{4.6}{\tiny$\pm$0.10} \\
SVHN & 97.6 & 49.0 & \spos{7.5}{\tiny$\pm$1.19} & 53.7 & \spos{5.7}{\tiny$\pm$0.83} & 74.3 & \sneg{--1.3}{\tiny$\pm$0.85} & 76.4 & \spos{0.4}{\tiny$\pm$0.65} & 59.7 & \spos{12.7}{\tiny$\pm$0.36} & 81.3 & \sneg{--1.8}{\tiny$\pm$0.75} \\
GTSRB & 99.1 & 43.2 & \spos{15.4}{\tiny$\pm$2.27} & 48.8 & \spos{13.5}{\tiny$\pm$2.00} & 77.6 & \spos{9.2}{\tiny$\pm$0.42} & 77.2 & \spos{9.7}{\tiny$\pm$0.60} & 55.7 & \spos{29.1}{\tiny$\pm$0.53} & 81.0 & \spos{7.6}{\tiny$\pm$0.38} \\
MNIST & 99.7 & 72.5 & \spos{12.5}{\tiny$\pm$0.61} & 79.0 & \spos{8.9}{\tiny$\pm$0.11} & 96.6 & \sneg{--0.5}{\tiny$\pm$0.08} & 94.8 & \sneg{--0.3}{\tiny$\pm$0.30} & 92.0 & \spos{2.3}{\tiny$\pm$0.33} & 95.4 & \sneg{--1.2}{\tiny$\pm$0.39} \\
DTD & 79.0 & 45.9 & \spos{6.3}{\tiny$\pm$0.32} & 48.4 & \spos{6.4}{\tiny$\pm$0.41} & 62.9 & \spos{6.5}{\tiny$\pm$0.27} & 60.6 & \spos{7.0}{\tiny$\pm$0.60} & 45.3 & \spos{18.8}{\tiny$\pm$0.29} & 63.0 & \spos{5.4}{\tiny$\pm$0.23} \\
Flowers102 & 90.4 & 66.1 & \spos{1.5}{\tiny$\pm$0.44} & 64.6 & \spos{3.5}{\tiny$\pm$0.34} & 77.6 & \spos{5.0}{\tiny$\pm$0.20} & 69.3 & \spos{10.0}{\tiny$\pm$0.41} & 53.9 & \spos{23.6}{\tiny$\pm$0.39} & 75.3 & \spos{4.3}{\tiny$\pm$0.19} \\
PCAM & 87.9 & 63.4 & \spos{6.2}{\tiny$\pm$1.96} & 67.0 & \spos{0.6}{\tiny$\pm$1.99} & 82.4 & \sneg{--7.6}{\tiny$\pm$1.55} & 84.5 & \sneg{--8.1}{\tiny$\pm$1.21} & 77.8 & \sneg{--3.1}{\tiny$\pm$0.68} & 80.8 & \sneg{--11.0}{\tiny$\pm$1.03} \\
FER2013 & 72.9 & 41.1 & \spos{4.6}{\tiny$\pm$0.56} & 42.4 & \spos{4.9}{\tiny$\pm$0.56} & 57.3 & \spos{6.4}{\tiny$\pm$0.20} & 58.3 & \spos{5.4}{\tiny$\pm$0.12} & 50.0 & \spos{11.7}{\tiny$\pm$0.13} & 63.1 & \spos{2.7}{\tiny$\pm$0.09} \\
OxfordPets & 91.3 & 87.2 & \spos{0.3}{\tiny$\pm$0.34} & 86.9 & \spos{0.5}{\tiny$\pm$0.30} & 87.5 & \spos{2.7}{\tiny$\pm$0.08} & 87.9 & \spos{2.4}{\tiny$\pm$0.06} & 76.2 & \spos{11.8}{\tiny$\pm$0.17} & 88.7 & \spos{1.6}{\tiny$\pm$0.11} \\
STL10 & 98.0 & 96.3 & \sneg{--0.2}{\tiny$\pm$0.10} & 95.8 & \spos{0.4}{\tiny$\pm$0.07} & 96.7 & \spos{0.5}{\tiny$\pm$0.10} & 95.8 & \spos{1.4}{\tiny$\pm$0.03} & 89.9 & \spos{6.5}{\tiny$\pm$0.13} & 96.8 & \spos{0.6}{\tiny$\pm$0.07} \\
CIFAR100 & 89.3 & 68.6 & \spos{1.9}{\tiny$\pm$0.28} & 68.5 & \spos{3.4}{\tiny$\pm$0.38} & 72.9 & \spos{5.2}{\tiny$\pm$0.07} & 65.8 & \spos{7.9}{\tiny$\pm$0.30} & 54.1 & \spos{18.3}{\tiny$\pm$0.41} & 68.5 & \spos{5.7}{\tiny$\pm$0.25} \\
CIFAR10 & 97.9 & 92.7 & \spos{0.3}{\tiny$\pm$0.38} & 92.5 & \spos{0.7}{\tiny$\pm$0.28} & 94.6 & \spos{1.0}{\tiny$\pm$0.11} & 93.3 & \spos{2.0}{\tiny$\pm$0.10} & 88.0 & \spos{6.3}{\tiny$\pm$0.05} & 93.8 & \spos{1.3}{\tiny$\pm$0.12} \\
Food101 & 89.1 & 78.8 & \spos{4.1}{\tiny$\pm$0.08} & 77.3 & \spos{5.6}{\tiny$\pm$0.15} & 76.1 & \spos{7.8}{\tiny$\pm$0.09} & 74.1 & \spos{10.4}{\tiny$\pm$0.16} & 43.0 & \spos{37.7}{\tiny$\pm$0.10} & 79.9 & \spos{5.2}{\tiny$\pm$0.10} \\
FashionMNIST & 95.3 & 71.9 & \spos{0.1}{\tiny$\pm$0.68} & 73.5 & \sneg{--0.2}{\tiny$\pm$0.71} & 86.6 & \spos{0.7}{\tiny$\pm$0.13} & 86.0 & \spos{1.4}{\tiny$\pm$0.15} & 79.7 & \spos{5.3}{\tiny$\pm$0.49} & 86.1 & \spos{0.7}{\tiny$\pm$0.16} \\
EMNIST & 99.8 & 34.7 & \spos{18.2}{\tiny$\pm$2.15} & 43.4 & \spos{20.0}{\tiny$\pm$2.26} & 88.1 & \spos{3.9}{\tiny$\pm$0.81} & 96.4 & \sneg{--0.2}{\tiny$\pm$0.34} & 93.9 & \spos{2.1}{\tiny$\pm$0.31} & 96.9 & \sneg{--0.2}{\tiny$\pm$0.28} \\
KMNIST & 98.6 & 9.4 & \spos{11.2}{\tiny$\pm$0.31} & 13.2 & \spos{14.0}{\tiny$\pm$0.51} & 60.9 & \spos{4.9}{\tiny$\pm$0.47} & 62.4 & \spos{8.3}{\tiny$\pm$0.27} & 45.0 & \spos{22.0}{\tiny$\pm$0.36} & 57.2 & \spos{11.5}{\tiny$\pm$0.68} \\
RenderedSST2 & 74.4 & 59.7 & \spos{0.5}{\tiny$\pm$0.64} & 60.2 & \spos{2.6}{\tiny$\pm$0.37} & 70.6 & \sneg{--6.5}{\tiny$\pm$0.22} & 69.6 & \spos{0.8}{\tiny$\pm$0.37} & 54.6 & \spos{16.3}{\tiny$\pm$0.29} & 73.0 & \sneg{--0.2}{\tiny$\pm$0.48} \\
\midrule
\rowcolor{grayrow}\textbf{Avg.} & 90.5 & 61.2 & \spos{5.6}{\tiny$\pm$0.22} & 63.1 & \spos{5.9}{\tiny$\pm$0.20} & 77.2 & \spos{3.6}{\tiny$\pm$0.13} & 76.1 & \spos{5.0}{\tiny$\pm$0.06} & 62.5 & \spos{16.5}{\tiny$\pm$0.05} & 78.5 & \spos{3.0}{\tiny$\pm$0.06} \\
\bottomrule
\end{tabular}}
\end{table}

%% file: figs/supp/pertask_ViT-B-16.tex
\begin{table}[p]
\centering
\caption{Per-task top-1 accuracy (\%) with ViT-B/16 of each base merge and its improvement after applying Merge++, reported as the mean $\pm$ standard deviation over generation seeds. Expert denotes the fine-tuned expert of each task.}
\label{tab:supp_pertask_vitb16}
\scriptsize\setlength{\tabcolsep}{3pt}
\resizebox{\linewidth}{!}{%
\begin{tabular}{l|c|cc|cc|cc|cc|cc|cc}
\toprule
 & & \multicolumn{2}{c|}{\textbf{TA} {\tiny\color{gray}[ICLR'23]}} & \multicolumn{2}{c|}{\textbf{TIES} {\tiny\color{gray}[NeurIPS'23]}} & \multicolumn{2}{c|}{\textbf{Iso-CTS} {\tiny\color{gray}[ICML'25]}} & \multicolumn{2}{c|}{\textbf{TSV-M} {\tiny\color{gray}[CVPR'25]}} & \multicolumn{2}{c|}{\textbf{WUDI} {\tiny\color{gray}[ICML'25]}} & \multicolumn{2}{c}{\textbf{SWUDI-A} {\tiny\color{gray}[arXiv'26]}} \\
\textbf{Task} & Expert & base & {\tiny\textbf{w/ Merge++}} & base & {\tiny\textbf{w/ Merge++}} & base & {\tiny\textbf{w/ Merge++}} & base & {\tiny\textbf{w/ Merge++}} & base & {\tiny\textbf{w/ Merge++}} & base & {\tiny\textbf{w/ Merge++}} \\
\midrule[\heavyrulewidth]
\multicolumn{14}{l}{\textbf{8 tasks}} \\
\midrule
SUN397 & 78.8 & 66.8 & \spos{5.2}{\tiny$\pm$0.16} & 68.3 & \spos{5.5}{\tiny$\pm$0.07} & 74.0 & \spos{2.3}{\tiny$\pm$0.08} & 71.3 & \spos{3.8}{\tiny$\pm$0.10} & 73.0 & \spos{2.5}{\tiny$\pm$0.06} & 74.0 & \spos{1.9}{\tiny$\pm$0.05} \\
Cars & 87.3 & 70.2 & \spos{3.1}{\tiny$\pm$0.87} & 73.8 & \spos{3.1}{\tiny$\pm$0.84} & 83.9 & \spos{0.5}{\tiny$\pm$0.50} & 81.6 & \spos{1.7}{\tiny$\pm$0.44} & 82.2 & \spos{0.6}{\tiny$\pm$0.39} & 82.2 & \spos{0.5}{\tiny$\pm$0.50} \\
RESISC45 & 97.0 & 76.3 & \spos{5.7}{\tiny$\pm$0.15} & 80.3 & \spos{5.8}{\tiny$\pm$0.29} & 93.4 & \spos{0.5}{\tiny$\pm$0.09} & 88.7 & \spos{3.0}{\tiny$\pm$0.21} & 89.6 & \spos{2.1}{\tiny$\pm$0.29} & 90.8 & \spos{1.8}{\tiny$\pm$0.12} \\
EuroSAT & 99.9 & 84.1 & \spos{5.2}{\tiny$\pm$0.45} & 91.0 & \spos{0.8}{\tiny$\pm$0.48} & 98.7 & \sneg{--2.5}{\tiny$\pm$0.08} & 97.7 & \sneg{--1.4}{\tiny$\pm$0.28} & 98.4 & \sneg{--1.4}{\tiny$\pm$0.04} & 98.9 & \sneg{--1.6}{\tiny$\pm$0.11} \\
SVHN & 97.8 & 82.9 & \sneg{--0.1}{\tiny$\pm$0.33} & 87.7 & \sneg{--2.6}{\tiny$\pm$0.33} & 90.6 & \sneg{--1.3}{\tiny$\pm$0.51} & 93.1 & \sneg{--2.3}{\tiny$\pm$0.35} & 94.4 & \sneg{--0.4}{\tiny$\pm$0.17} & 94.7 & \sneg{--0.5}{\tiny$\pm$0.17} \\
GTSRB & 99.0 & 73.6 & \spos{12.9}{\tiny$\pm$0.20} & 73.4 & \spos{13.3}{\tiny$\pm$0.14} & 97.1 & \spos{0.2}{\tiny$\pm$0.05} & 94.0 & \spos{1.9}{\tiny$\pm$0.06} & 96.4 & \spos{1.1}{\tiny$\pm$0.10} & 96.6 & \spos{0.8}{\tiny$\pm$0.08} \\
MNIST & 99.8 & 98.0 & \sneg{--0.1}{\tiny$\pm$0.22} & 98.6 & \sneg{--0.1}{\tiny$\pm$0.18} & 99.2 & \spos{0.1}{\tiny$\pm$0.02} & 99.3 & \szero{0.0}{\tiny$\pm$0.03} & 99.4 & \spos{0.1}{\tiny$\pm$0.03} & 99.5 & \szero{0.0}{\tiny$\pm$0.02} \\
DTD & 82.1 & 53.0 & \spos{10.5}{\tiny$\pm$0.35} & 57.7 & \spos{10.1}{\tiny$\pm$0.33} & 76.5 & \spos{2.0}{\tiny$\pm$0.27} & 72.0 & \spos{4.8}{\tiny$\pm$0.25} & 73.7 & \spos{3.6}{\tiny$\pm$0.13} & 76.3 & \spos{1.7}{\tiny$\pm$0.19} \\
\midrule
\rowcolor{grayrow}\textbf{Avg.} & 92.7 & 75.6 & \spos{5.3}{\tiny$\pm$0.18} & 78.8 & \spos{4.5}{\tiny$\pm$0.11} & 89.2 & \spos{0.2}{\tiny$\pm$0.11} & 87.2 & \spos{1.4}{\tiny$\pm$0.09} & 88.4 & \spos{1.0}{\tiny$\pm$0.06} & 89.1 & \spos{0.6}{\tiny$\pm$0.07} \\
\midrule[\heavyrulewidth]
\multicolumn{14}{l}{\textbf{14 tasks}} \\
\midrule
SUN397 & 78.8 & 67.3 & \spos{2.8}{\tiny$\pm$0.02} & 68.7 & \spos{2.8}{\tiny$\pm$0.00} & 71.4 & \spos{2.9}{\tiny$\pm$0.15} & 69.5 & \spos{3.6}{\tiny$\pm$0.17} & 68.4 & \spos{4.8}{\tiny$\pm$0.19} & 72.2 & \spos{1.7}{\tiny$\pm$0.27} \\
Cars & 87.3 & 65.4 & \spos{2.4}{\tiny$\pm$0.47} & 68.5 & \spos{2.1}{\tiny$\pm$0.67} & 76.8 & \spos{3.1}{\tiny$\pm$0.55} & 73.9 & \spos{5.0}{\tiny$\pm$0.33} & 73.0 & \spos{5.1}{\tiny$\pm$0.64} & 77.3 & \spos{1.8}{\tiny$\pm$0.24} \\
RESISC45 & 97.0 & 71.5 & \spos{5.1}{\tiny$\pm$0.17} & 74.6 & \spos{4.9}{\tiny$\pm$0.15} & 90.9 & \spos{1.1}{\tiny$\pm$0.17} & 86.0 & \spos{3.0}{\tiny$\pm$0.21} & 84.1 & \spos{4.6}{\tiny$\pm$0.23} & 88.2 & \spos{1.4}{\tiny$\pm$0.25} \\
EuroSAT & 99.9 & 75.4 & \spos{6.7}{\tiny$\pm$0.42} & 83.3 & \spos{1.7}{\tiny$\pm$0.37} & 97.6 & \sneg{--2.9}{\tiny$\pm$0.13} & 96.1 & \sneg{--1.1}{\tiny$\pm$0.15} & 94.7 & \sneg{--0.3}{\tiny$\pm$0.15} & 97.4 & \sneg{--1.4}{\tiny$\pm$0.40} \\
SVHN & 97.8 & 70.5 & \spos{4.2}{\tiny$\pm$0.49} & 75.7 & \spos{1.3}{\tiny$\pm$0.49} & 89.2 & \sneg{--1.4}{\tiny$\pm$0.44} & 90.6 & \sneg{--2.2}{\tiny$\pm$0.39} & 90.4 & \spos{0.7}{\tiny$\pm$0.36} & 92.6 & \sneg{--0.4}{\tiny$\pm$0.20} \\
GTSRB & 99.0 & 61.4 & \spos{15.2}{\tiny$\pm$0.37} & 63.8 & \spos{14.4}{\tiny$\pm$0.27} & 94.3 & \spos{1.0}{\tiny$\pm$0.07} & 90.5 & \spos{2.7}{\tiny$\pm$0.09} & 90.1 & \spos{4.5}{\tiny$\pm$0.05} & 93.5 & \spos{1.8}{\tiny$\pm$0.09} \\
MNIST & 99.8 & 93.7 & \spos{1.2}{\tiny$\pm$0.32} & 96.0 & \spos{0.4}{\tiny$\pm$0.22} & 99.0 & \spos{0.2}{\tiny$\pm$0.02} & 99.2 & \szero{0.0}{\tiny$\pm$0.03} & 99.2 & \spos{0.1}{\tiny$\pm$0.03} & 99.4 & \szero{0.0}{\tiny$\pm$0.01} \\
DTD & 82.1 & 48.4 & \spos{8.8}{\tiny$\pm$0.79} & 50.5 & \spos{9.7}{\tiny$\pm$0.17} & 71.0 & \spos{4.1}{\tiny$\pm$0.45} & 65.2 & \spos{7.5}{\tiny$\pm$0.15} & 62.7 & \spos{9.6}{\tiny$\pm$0.06} & 69.6 & \spos{4.2}{\tiny$\pm$0.47} \\
Flowers102 & 93.7 & 71.2 & \spos{2.9}{\tiny$\pm$0.50} & 69.9 & \spos{3.5}{\tiny$\pm$0.50} & 87.9 & \spos{3.1}{\tiny$\pm$0.08} & 81.5 & \spos{4.8}{\tiny$\pm$0.10} & 78.6 & \spos{6.6}{\tiny$\pm$0.29} & 85.9 & \spos{1.5}{\tiny$\pm$0.06} \\
PCAM & 89.2 & 57.5 & \spos{14.5}{\tiny$\pm$0.40} & 60.6 & \spos{11.2}{\tiny$\pm$0.47} & 76.9 & \sneg{--9.3}{\tiny$\pm$1.70} & 85.3 & \sneg{--13.3}{\tiny$\pm$1.38} & 85.3 & \sneg{--14.8}{\tiny$\pm$1.31} & 84.6 & \sneg{--15.2}{\tiny$\pm$1.14} \\
FER2013 & 74.7 & 44.8 & \spos{4.5}{\tiny$\pm$0.43} & 47.3 & \spos{5.1}{\tiny$\pm$0.18} & 61.9 & \spos{6.7}{\tiny$\pm$0.30} & 65.3 & \spos{3.9}{\tiny$\pm$0.07} & 60.4 & \spos{8.9}{\tiny$\pm$0.18} & 68.6 & \spos{1.9}{\tiny$\pm$0.18} \\
OxfordPets & 94.5 & 90.6 & \spos{1.7}{\tiny$\pm$0.21} & 90.9 & \spos{1.2}{\tiny$\pm$0.15} & 92.9 & \spos{1.0}{\tiny$\pm$0.07} & 93.9 & \spos{0.4}{\tiny$\pm$0.14} & 93.7 & \spos{0.5}{\tiny$\pm$0.18} & 94.2 & \spos{0.1}{\tiny$\pm$0.19} \\
STL10 & 98.6 & 97.5 & \spos{0.5}{\tiny$\pm$0.10} & 97.3 & \spos{0.6}{\tiny$\pm$0.15} & 98.2 & \spos{0.1}{\tiny$\pm$0.04} & 98.0 & \spos{0.5}{\tiny$\pm$0.04} & 97.7 & \spos{0.6}{\tiny$\pm$0.04} & 98.3 & \spos{0.2}{\tiny$\pm$0.10} \\
CIFAR100 & 90.7 & 72.7 & \spos{3.4}{\tiny$\pm$0.43} & 73.2 & \spos{4.2}{\tiny$\pm$0.22} & 80.3 & \spos{1.8}{\tiny$\pm$0.07} & 76.2 & \spos{3.9}{\tiny$\pm$0.35} & 75.4 & \spos{4.0}{\tiny$\pm$0.31} & 78.1 & \spos{2.3}{\tiny$\pm$0.34} \\
\midrule
\rowcolor{grayrow}\textbf{Avg.} & 91.7 & 70.6 & \spos{5.3}{\tiny$\pm$0.10} & 72.9 & \spos{4.5}{\tiny$\pm$0.08} & 84.9 & \spos{0.8}{\tiny$\pm$0.07} & 83.7 & \spos{1.3}{\tiny$\pm$0.12} & 82.4 & \spos{2.5}{\tiny$\pm$0.12} & 85.7 & \szero{0.0}{\tiny$\pm$0.12} \\
\midrule[\heavyrulewidth]
\multicolumn{14}{l}{\textbf{20 tasks}} \\
\midrule
SUN397 & 78.8 & 67.4 & \spos{1.8}{\tiny$\pm$0.04} & 68.7 & \spos{1.6}{\tiny$\pm$0.14} & 70.1 & \spos{3.1}{\tiny$\pm$0.11} & 67.9 & \spos{4.1}{\tiny$\pm$0.15} & 56.0 & \spos{14.6}{\tiny$\pm$0.15} & 71.0 & \spos{1.6}{\tiny$\pm$0.08} \\
Cars & 87.3 & 64.9 & \spos{1.1}{\tiny$\pm$0.06} & 65.5 & \spos{1.9}{\tiny$\pm$0.42} & 66.5 & \spos{8.5}{\tiny$\pm$0.79} & 61.9 & \spos{11.5}{\tiny$\pm$1.04} & 49.3 & \spos{20.6}{\tiny$\pm$1.11} & 70.4 & \spos{5.0}{\tiny$\pm$0.48} \\
RESISC45 & 97.0 & 69.8 & \spos{3.3}{\tiny$\pm$0.04} & 72.6 & \spos{2.9}{\tiny$\pm$0.13} & 87.6 & \spos{3.3}{\tiny$\pm$0.10} & 80.5 & \spos{6.1}{\tiny$\pm$0.02} & 67.2 & \spos{18.0}{\tiny$\pm$0.15} & 85.7 & \spos{2.1}{\tiny$\pm$0.11} \\
EuroSAT & 99.9 & 63.2 & \spos{12.0}{\tiny$\pm$0.58} & 71.4 & \spos{7.7}{\tiny$\pm$0.79} & 95.9 & \sneg{--1.6}{\tiny$\pm$0.58} & 93.8 & \sneg{--0.2}{\tiny$\pm$0.35} & 75.9 & \spos{12.5}{\tiny$\pm$0.42} & 95.2 & \sneg{--1.0}{\tiny$\pm$0.43} \\
SVHN & 97.8 & 65.3 & \spos{5.1}{\tiny$\pm$0.67} & 70.2 & \spos{3.2}{\tiny$\pm$0.49} & 86.9 & \sneg{--1.3}{\tiny$\pm$0.16} & 87.1 & \sneg{--1.5}{\tiny$\pm$0.17} & 79.1 & \spos{5.7}{\tiny$\pm$0.15} & 88.7 & \sneg{--0.4}{\tiny$\pm$0.21} \\
GTSRB & 99.0 & 54.0 & \spos{16.3}{\tiny$\pm$0.97} & 56.5 & \spos{16.5}{\tiny$\pm$0.81} & 88.6 & \spos{4.1}{\tiny$\pm$0.27} & 84.9 & \spos{6.0}{\tiny$\pm$0.44} & 69.0 & \spos{20.5}{\tiny$\pm$0.31} & 88.5 & \spos{3.8}{\tiny$\pm$0.30} \\
MNIST & 99.8 & 88.5 & \spos{3.3}{\tiny$\pm$0.66} & 92.2 & \spos{1.3}{\tiny$\pm$0.50} & 98.1 & \spos{0.1}{\tiny$\pm$0.06} & 97.5 & \spos{0.3}{\tiny$\pm$0.06} & 96.3 & \spos{1.3}{\tiny$\pm$0.10} & 98.0 & \sneg{--0.1}{\tiny$\pm$0.14} \\
DTD & 82.1 & 47.2 & \spos{7.0}{\tiny$\pm$0.40} & 49.3 & \spos{6.9}{\tiny$\pm$0.41} & 66.8 & \spos{5.7}{\tiny$\pm$0.57} & 61.5 & \spos{9.6}{\tiny$\pm$0.37} & 49.7 & \spos{19.5}{\tiny$\pm$0.19} & 65.6 & \spos{6.3}{\tiny$\pm$0.15} \\
Flowers102 & 93.7 & 71.6 & \spos{0.8}{\tiny$\pm$0.53} & 71.0 & \spos{1.8}{\tiny$\pm$0.53} & 84.3 & \spos{5.0}{\tiny$\pm$0.30} & 77.6 & \spos{6.4}{\tiny$\pm$0.29} & 63.8 & \spos{18.1}{\tiny$\pm$0.21} & 82.0 & \spos{2.7}{\tiny$\pm$0.19} \\
PCAM & 89.2 & 59.3 & \spos{8.2}{\tiny$\pm$2.39} & 59.1 & \spos{9.8}{\tiny$\pm$1.82} & 74.0 & \sneg{--10.1}{\tiny$\pm$1.21} & 84.5 & \sneg{--11.4}{\tiny$\pm$0.91} & 79.3 & \sneg{--12.1}{\tiny$\pm$1.21} & 85.0 & \sneg{--16.6}{\tiny$\pm$0.89} \\
FER2013 & 74.7 & 41.3 & \spos{4.8}{\tiny$\pm$0.87} & 43.2 & \spos{4.8}{\tiny$\pm$0.68} & 59.0 & \spos{6.4}{\tiny$\pm$0.04} & 60.4 & \spos{6.2}{\tiny$\pm$0.23} & 44.1 & \spos{20.1}{\tiny$\pm$0.22} & 64.9 & \spos{3.4}{\tiny$\pm$0.05} \\
OxfordPets & 94.5 & 90.1 & \spos{0.9}{\tiny$\pm$0.70} & 90.5 & \spos{0.6}{\tiny$\pm$0.58} & 92.6 & \spos{0.8}{\tiny$\pm$0.23} & 93.6 & \spos{0.4}{\tiny$\pm$0.14} & 88.8 & \spos{4.5}{\tiny$\pm$0.15} & 93.5 & \spos{0.6}{\tiny$\pm$0.09} \\
STL10 & 98.6 & 97.9 & \sneg{--0.2}{\tiny$\pm$0.16} & 97.6 & \sneg{--0.1}{\tiny$\pm$0.20} & 97.6 & \spos{0.3}{\tiny$\pm$0.11} & 97.4 & \spos{0.8}{\tiny$\pm$0.08} & 94.5 & \spos{2.7}{\tiny$\pm$0.17} & 98.0 & \spos{0.1}{\tiny$\pm$0.13} \\
CIFAR100 & 90.7 & 73.8 & \spos{1.0}{\tiny$\pm$0.42} & 73.8 & \spos{2.2}{\tiny$\pm$0.35} & 77.0 & \spos{3.4}{\tiny$\pm$0.14} & 72.3 & \spos{5.3}{\tiny$\pm$0.27} & 62.2 & \spos{12.3}{\tiny$\pm$0.19} & 75.6 & \spos{2.9}{\tiny$\pm$0.14} \\
CIFAR10 & 98.5 & 94.3 & \szero{0.0}{\tiny$\pm$0.09} & 94.5 & \spos{0.3}{\tiny$\pm$0.11} & 95.8 & \spos{0.6}{\tiny$\pm$0.13} & 95.5 & \spos{0.8}{\tiny$\pm$0.13} & 90.2 & \spos{5.3}{\tiny$\pm$0.08} & 95.9 & \spos{0.3}{\tiny$\pm$0.06} \\
Food101 & 92.9 & 86.3 & \spos{1.8}{\tiny$\pm$0.10} & 86.1 & \spos{2.4}{\tiny$\pm$0.05} & 84.9 & \spos{4.2}{\tiny$\pm$0.07} & 84.4 & \spos{5.2}{\tiny$\pm$0.07} & 67.4 & \spos{20.1}{\tiny$\pm$0.07} & 87.6 & \spos{2.5}{\tiny$\pm$0.05} \\
FashionMNIST & 95.7 & 74.6 & \sneg{--1.3}{\tiny$\pm$0.47} & 77.7 & \sneg{--1.6}{\tiny$\pm$0.19} & 89.6 & \sneg{--1.2}{\tiny$\pm$0.16} & 89.2 & \sneg{--1.1}{\tiny$\pm$0.28} & 81.6 & \spos{3.8}{\tiny$\pm$0.54} & 88.8 & \sneg{--1.5}{\tiny$\pm$0.46} \\
EMNIST & 99.8 & 34.4 & \spos{12.8}{\tiny$\pm$0.89} & 44.6 & \spos{15.3}{\tiny$\pm$1.48} & 88.7 & \spos{2.3}{\tiny$\pm$0.26} & 96.4 & \sneg{--0.6}{\tiny$\pm$0.27} & 92.3 & \spos{2.3}{\tiny$\pm$0.30} & 96.2 & \sneg{--0.5}{\tiny$\pm$0.23} \\
KMNIST & 98.8 & 11.0 & \spos{7.3}{\tiny$\pm$0.70} & 14.4 & \spos{8.9}{\tiny$\pm$0.79} & 52.3 & \spos{14.6}{\tiny$\pm$2.40} & 42.7 & \spos{21.5}{\tiny$\pm$2.06} & 27.1 & \spos{24.1}{\tiny$\pm$2.61} & 37.5 & \spos{19.9}{\tiny$\pm$1.95} \\
RenderedSST2 & 79.2 & 60.7 & \sneg{--1.6}{\tiny$\pm$0.51} & 61.2 & \spos{1.8}{\tiny$\pm$0.38} & 75.7 & \spos{1.9}{\tiny$\pm$0.16} & 74.9 & \spos{2.4}{\tiny$\pm$0.29} & 62.7 & \spos{13.6}{\tiny$\pm$0.31} & 77.4 & \spos{2.0}{\tiny$\pm$0.03} \\
\midrule
\rowcolor{grayrow}\textbf{Avg.} & 92.4 & 65.8 & \spos{4.2}{\tiny$\pm$0.08} & 68.0 & \spos{4.4}{\tiny$\pm$0.06} & 81.6 & \spos{2.5}{\tiny$\pm$0.08} & 80.2 & \spos{3.6}{\tiny$\pm$0.06} & 69.8 & \spos{11.4}{\tiny$\pm$0.10} & 82.3 & \spos{1.7}{\tiny$\pm$0.04} \\
\bottomrule
\end{tabular}}
\end{table}

%% file: figs/supp/pertask_ViT-L-14.tex
\begin{table}[p]
\centering
\caption{Per-task top-1 accuracy (\%) with ViT-L/14 of each base merge and its improvement after applying Merge++, reported as the mean $\pm$ standard deviation over generation seeds. Expert denotes the fine-tuned expert of each task.}
\label{tab:supp_pertask_vitl14}
\scriptsize\setlength{\tabcolsep}{3pt}
\resizebox{\linewidth}{!}{%
\begin{tabular}{l|c|cc|cc|cc|cc|cc|cc}
\toprule
 & & \multicolumn{2}{c|}{\textbf{TA} {\tiny\color{gray}[ICLR'23]}} & \multicolumn{2}{c|}{\textbf{TIES} {\tiny\color{gray}[NeurIPS'23]}} & \multicolumn{2}{c|}{\textbf{Iso-CTS} {\tiny\color{gray}[ICML'25]}} & \multicolumn{2}{c|}{\textbf{TSV-M} {\tiny\color{gray}[CVPR'25]}} & \multicolumn{2}{c|}{\textbf{WUDI} {\tiny\color{gray}[ICML'25]}} & \multicolumn{2}{c}{\textbf{SWUDI-A} {\tiny\color{gray}[arXiv'26]}} \\
\textbf{Task} & Expert & base & {\tiny\textbf{w/ Merge++}} & base & {\tiny\textbf{w/ Merge++}} & base & {\tiny\textbf{w/ Merge++}} & base & {\tiny\textbf{w/ Merge++}} & base & {\tiny\textbf{w/ Merge++}} & base & {\tiny\textbf{w/ Merge++}} \\
\midrule[\heavyrulewidth]
\multicolumn{14}{l}{\textbf{8 tasks}} \\
\midrule
SUN397 & 82.1 & 73.6 & \spos{3.4}{\tiny$\pm$0.03} & 75.6 & \spos{3.8}{\tiny$\pm$0.04} & 80.5 & \spos{0.8}{\tiny$\pm$0.04} & 78.1 & \spos{1.8}{\tiny$\pm$0.05} & 78.9 & \spos{1.4}{\tiny$\pm$0.03} & 79.8 & \spos{0.8}{\tiny$\pm$0.06} \\
Cars & 92.0 & 82.1 & \spos{2.8}{\tiny$\pm$0.36} & 84.6 & \spos{3.2}{\tiny$\pm$0.12} & 92.0 & \szero{0.0}{\tiny$\pm$0.21} & 89.7 & \spos{0.6}{\tiny$\pm$0.14} & 90.6 & \spos{0.1}{\tiny$\pm$0.16} & 90.4 & \szero{0.0}{\tiny$\pm$0.16} \\
RESISC45 & 97.3 & 87.1 & \spos{5.3}{\tiny$\pm$0.18} & 88.9 & \spos{4.9}{\tiny$\pm$0.10} & 95.8 & \spos{0.8}{\tiny$\pm$0.06} & 93.3 & \spos{2.3}{\tiny$\pm$0.11} & 94.0 & \spos{1.6}{\tiny$\pm$0.13} & 94.5 & \spos{1.2}{\tiny$\pm$0.08} \\
EuroSAT & 99.9 & 93.2 & \spos{2.6}{\tiny$\pm$0.10} & 94.9 & \spos{2.1}{\tiny$\pm$0.12} & 99.5 & \szero{0.0}{\tiny$\pm$0.08} & 99.0 & \sneg{--0.1}{\tiny$\pm$0.06} & 99.4 & \szero{0.0}{\tiny$\pm$0.04} & 99.6 & \sneg{--0.1}{\tiny$\pm$0.04} \\
SVHN & 98.1 & 86.7 & \spos{3.1}{\tiny$\pm$0.15} & 89.6 & \spos{1.9}{\tiny$\pm$0.19} & 95.2 & \sneg{--0.2}{\tiny$\pm$0.06} & 94.8 & \szero{0.0}{\tiny$\pm$0.06} & 95.9 & \spos{0.3}{\tiny$\pm$0.02} & 96.2 & \sneg{--0.2}{\tiny$\pm$0.05} \\
GTSRB & 99.1 & 85.6 & \spos{8.9}{\tiny$\pm$0.16} & 83.2 & \spos{11.0}{\tiny$\pm$0.09} & 98.6 & \spos{0.1}{\tiny$\pm$0.05} & 96.3 & \spos{1.6}{\tiny$\pm$0.07} & 98.3 & \spos{0.4}{\tiny$\pm$0.01} & 98.4 & \spos{0.3}{\tiny$\pm$0.03} \\
MNIST & 99.8 & 99.1 & \spos{0.2}{\tiny$\pm$0.02} & 99.2 & \spos{0.2}{\tiny$\pm$0.03} & 99.6 & \szero{0.0}{\tiny$\pm$0.03} & 99.5 & \spos{0.1}{\tiny$\pm$0.04} & 99.6 & \szero{0.0}{\tiny$\pm$0.01} & 99.7 & \szero{0.0}{\tiny$\pm$0.01} \\
DTD & 84.4 & 64.0 & \spos{8.1}{\tiny$\pm$0.21} & 67.6 & \spos{7.5}{\tiny$\pm$0.27} & 82.4 & \spos{1.2}{\tiny$\pm$0.16} & 78.9 & \spos{2.3}{\tiny$\pm$0.15} & 80.1 & \spos{1.2}{\tiny$\pm$0.29} & 81.6 & \spos{0.9}{\tiny$\pm$0.22} \\
\midrule
\rowcolor{grayrow}\textbf{Avg.} & 94.1 & 83.9 & \spos{4.3}{\tiny$\pm$0.09} & 85.5 & \spos{4.3}{\tiny$\pm$0.06} & 92.9 & \spos{0.3}{\tiny$\pm$0.02} & 91.2 & \spos{1.1}{\tiny$\pm$0.01} & 92.1 & \spos{0.6}{\tiny$\pm$0.08} & 92.5 & \spos{0.4}{\tiny$\pm$0.05} \\
\midrule[\heavyrulewidth]
\multicolumn{14}{l}{\textbf{14 tasks}} \\
\midrule
SUN397 & 82.1 & 71.8 & \spos{2.6}{\tiny$\pm$0.05} & 73.8 & \spos{2.6}{\tiny$\pm$0.05} & 78.4 & \spos{1.8}{\tiny$\pm$0.03} & 75.7 & \spos{2.3}{\tiny$\pm$0.07} & 76.7 & \spos{2.1}{\tiny$\pm$0.06} & 77.9 & \spos{1.1}{\tiny$\pm$0.05} \\
Cars & 92.0 & 78.3 & \spos{2.2}{\tiny$\pm$0.17} & 79.6 & \spos{2.9}{\tiny$\pm$0.09} & 88.8 & \spos{0.7}{\tiny$\pm$0.12} & 85.9 & \spos{2.6}{\tiny$\pm$0.27} & 87.9 & \spos{1.7}{\tiny$\pm$0.11} & 88.2 & \spos{1.0}{\tiny$\pm$0.13} \\
RESISC45 & 97.3 & 81.9 & \spos{5.9}{\tiny$\pm$0.26} & 84.1 & \spos{5.7}{\tiny$\pm$0.08} & 94.6 & \spos{1.2}{\tiny$\pm$0.12} & 92.4 & \spos{2.4}{\tiny$\pm$0.06} & 92.1 & \spos{2.7}{\tiny$\pm$0.04} & 93.4 & \spos{1.3}{\tiny$\pm$0.13} \\
EuroSAT & 99.9 & 85.7 & \spos{4.4}{\tiny$\pm$0.27} & 87.7 & \spos{4.2}{\tiny$\pm$0.27} & 99.2 & \szero{0.0}{\tiny$\pm$0.06} & 98.2 & \sneg{--0.1}{\tiny$\pm$0.15} & 98.1 & \szero{0.0}{\tiny$\pm$0.30} & 99.2 & \sneg{--0.3}{\tiny$\pm$0.10} \\
SVHN & 98.1 & 79.4 & \spos{5.7}{\tiny$\pm$0.50} & 81.7 & \spos{5.1}{\tiny$\pm$0.52} & 93.9 & \spos{0.3}{\tiny$\pm$0.01} & 93.6 & \spos{0.2}{\tiny$\pm$0.13} & 93.7 & \spos{0.9}{\tiny$\pm$0.08} & 94.4 & \spos{0.1}{\tiny$\pm$0.08} \\
GTSRB & 99.1 & 72.1 & \spos{16.4}{\tiny$\pm$0.34} & 70.6 & \spos{18.5}{\tiny$\pm$0.52} & 97.6 & \spos{0.6}{\tiny$\pm$0.04} & 94.3 & \spos{2.5}{\tiny$\pm$0.06} & 96.1 & \spos{1.8}{\tiny$\pm$0.14} & 97.1 & \spos{0.8}{\tiny$\pm$0.06} \\
MNIST & 99.8 & 97.5 & \spos{0.9}{\tiny$\pm$0.07} & 98.0 & \spos{0.6}{\tiny$\pm$0.09} & 99.5 & \szero{0.0}{\tiny$\pm$0.03} & 99.5 & \szero{0.0}{\tiny$\pm$0.03} & 99.6 & \szero{0.0}{\tiny$\pm$0.03} & 99.6 & \szero{0.0}{\tiny$\pm$0.03} \\
DTD & 84.4 & 60.6 & \spos{5.9}{\tiny$\pm$0.11} & 61.4 & \spos{6.7}{\tiny$\pm$0.23} & 79.9 & \spos{2.1}{\tiny$\pm$0.16} & 74.2 & \spos{4.4}{\tiny$\pm$0.34} & 75.1 & \spos{4.1}{\tiny$\pm$0.25} & 78.5 & \spos{1.6}{\tiny$\pm$0.11} \\
Flowers102 & 98.0 & 78.5 & \spos{5.9}{\tiny$\pm$0.17} & 66.4 & \spos{6.1}{\tiny$\pm$0.35} & 93.2 & \spos{3.2}{\tiny$\pm$0.15} & 87.5 & \spos{5.2}{\tiny$\pm$0.18} & 90.4 & \spos{2.9}{\tiny$\pm$0.05} & 93.2 & \spos{0.9}{\tiny$\pm$0.05} \\
PCAM & 90.3 & 81.0 & \sneg{--12.5}{\tiny$\pm$0.65} & 79.0 & \sneg{--10.8}{\tiny$\pm$0.47} & 78.8 & \sneg{--2.6}{\tiny$\pm$0.69} & 83.9 & \sneg{--5.4}{\tiny$\pm$0.71} & 84.2 & \sneg{--4.1}{\tiny$\pm$0.60} & 84.7 & \sneg{--3.7}{\tiny$\pm$0.59} \\
FER2013 & 77.0 & 47.0 & \spos{5.7}{\tiny$\pm$0.23} & 50.0 & \spos{6.8}{\tiny$\pm$0.47} & 68.2 & \spos{5.7}{\tiny$\pm$0.14} & 69.0 & \spos{4.0}{\tiny$\pm$0.05} & 69.5 & \spos{4.6}{\tiny$\pm$0.05} & 73.0 & \spos{2.0}{\tiny$\pm$0.04} \\
OxfordPets & 95.6 & 95.3 & \sneg{--0.2}{\tiny$\pm$0.20} & 94.7 & \spos{0.1}{\tiny$\pm$0.31} & 95.6 & \spos{0.2}{\tiny$\pm$0.16} & 96.2 & \szero{0.0}{\tiny$\pm$0.05} & 96.4 & \sneg{--0.2}{\tiny$\pm$0.09} & 96.4 & \sneg{--0.1}{\tiny$\pm$0.11} \\
STL10 & 99.5 & 99.1 & \spos{0.1}{\tiny$\pm$0.05} & 98.8 & \spos{0.5}{\tiny$\pm$0.03} & 99.4 & \spos{0.2}{\tiny$\pm$0.03} & 99.5 & \szero{0.0}{\tiny$\pm$0.05} & 99.5 & \szero{0.0}{\tiny$\pm$0.01} & 99.5 & \sneg{--0.1}{\tiny$\pm$0.02} \\
CIFAR100 & 93.3 & 82.6 & \spos{2.0}{\tiny$\pm$0.14} & 82.5 & \spos{3.1}{\tiny$\pm$0.10} & 87.6 & \spos{1.5}{\tiny$\pm$0.02} & 85.6 & \spos{2.3}{\tiny$\pm$0.03} & 85.8 & \spos{2.1}{\tiny$\pm$0.09} & 87.2 & \spos{1.2}{\tiny$\pm$0.05} \\
\midrule
\rowcolor{grayrow}\textbf{Avg.} & 93.3 & 79.3 & \spos{3.2}{\tiny$\pm$0.08} & 79.2 & \spos{3.7}{\tiny$\pm$0.06} & 89.6 & \spos{1.1}{\tiny$\pm$0.03} & 88.3 & \spos{1.5}{\tiny$\pm$0.03} & 88.9 & \spos{1.3}{\tiny$\pm$0.02} & 90.2 & \spos{0.4}{\tiny$\pm$0.05} \\
\midrule[\heavyrulewidth]
\multicolumn{14}{l}{\textbf{20 tasks}} \\
\midrule
SUN397 & 82.1 & 70.8 & \spos{2.8}{\tiny$\pm$0.06} & 72.3 & \spos{2.5}{\tiny$\pm$0.11} & 77.9 & \spos{1.7}{\tiny$\pm$0.06} & 74.4 & \spos{2.6}{\tiny$\pm$0.08} & 72.8 & \spos{4.8}{\tiny$\pm$0.09} & 76.6 & \spos{1.3}{\tiny$\pm$0.11} \\
Cars & 92.0 & 73.9 & \spos{3.5}{\tiny$\pm$0.89} & 75.8 & \spos{2.8}{\tiny$\pm$0.79} & 85.4 & \spos{2.9}{\tiny$\pm$0.78} & 80.4 & \spos{6.4}{\tiny$\pm$0.74} & 75.9 & \spos{10.1}{\tiny$\pm$0.86} & 85.3 & \spos{2.4}{\tiny$\pm$0.52} \\
RESISC45 & 97.3 & 76.5 & \spos{8.7}{\tiny$\pm$0.37} & 79.7 & \spos{6.8}{\tiny$\pm$0.34} & 93.9 & \spos{1.7}{\tiny$\pm$0.08} & 90.7 & \spos{3.5}{\tiny$\pm$0.03} & 84.6 & \spos{9.1}{\tiny$\pm$0.08} & 92.2 & \spos{1.8}{\tiny$\pm$0.10} \\
EuroSAT & 99.9 & 74.8 & \spos{11.2}{\tiny$\pm$0.19} & 78.6 & \spos{8.7}{\tiny$\pm$0.40} & 98.4 & \spos{0.2}{\tiny$\pm$0.11} & 96.1 & \spos{1.6}{\tiny$\pm$0.09} & 82.5 & \spos{13.3}{\tiny$\pm$0.13} & 98.4 & \sneg{--0.1}{\tiny$\pm$0.02} \\
SVHN & 98.1 & 76.2 & \spos{4.9}{\tiny$\pm$0.35} & 78.0 & \spos{4.4}{\tiny$\pm$0.40} & 90.0 & \spos{1.3}{\tiny$\pm$0.18} & 90.0 & \spos{1.0}{\tiny$\pm$0.11} & 84.6 & \spos{4.8}{\tiny$\pm$0.26} & 91.7 & \spos{0.1}{\tiny$\pm$0.13} \\
GTSRB & 99.1 & 66.1 & \spos{19.4}{\tiny$\pm$1.13} & 64.1 & \spos{21.4}{\tiny$\pm$0.98} & 95.7 & \spos{1.7}{\tiny$\pm$0.09} & 90.8 & \spos{4.8}{\tiny$\pm$0.16} & 84.8 & \spos{11.1}{\tiny$\pm$0.31} & 94.3 & \spos{2.4}{\tiny$\pm$0.08} \\
MNIST & 99.8 & 96.2 & \spos{0.4}{\tiny$\pm$0.31} & 96.5 & \spos{0.2}{\tiny$\pm$0.24} & 98.7 & \sneg{--0.2}{\tiny$\pm$0.08} & 97.3 & \spos{0.4}{\tiny$\pm$0.08} & 97.0 & \spos{0.9}{\tiny$\pm$0.04} & 97.5 & \spos{0.3}{\tiny$\pm$0.07} \\
DTD & 84.4 & 58.0 & \spos{5.1}{\tiny$\pm$0.60} & 59.7 & \spos{5.0}{\tiny$\pm$0.09} & 78.2 & \spos{2.1}{\tiny$\pm$0.08} & 71.3 & \spos{4.6}{\tiny$\pm$0.19} & 66.1 & \spos{10.5}{\tiny$\pm$0.19} & 76.1 & \spos{1.8}{\tiny$\pm$0.18} \\
Flowers102 & 98.0 & 75.8 & \spos{6.9}{\tiny$\pm$0.27} & 69.2 & \spos{5.6}{\tiny$\pm$0.41} & 95.4 & \spos{1.3}{\tiny$\pm$0.07} & 85.6 & \spos{6.1}{\tiny$\pm$0.29} & 80.1 & \spos{10.8}{\tiny$\pm$0.39} & 91.3 & \spos{1.5}{\tiny$\pm$0.47} \\
PCAM & 90.3 & 75.0 & \sneg{--9.1}{\tiny$\pm$1.35} & 75.8 & \sneg{--10.4}{\tiny$\pm$1.22} & 79.2 & \sneg{--5.1}{\tiny$\pm$0.38} & 85.0 & \sneg{--7.0}{\tiny$\pm$0.28} & 81.1 & \sneg{--5.7}{\tiny$\pm$0.39} & 85.6 & \sneg{--6.2}{\tiny$\pm$0.19} \\
FER2013 & 77.0 & 42.7 & \spos{6.4}{\tiny$\pm$0.21} & 44.9 & \spos{6.3}{\tiny$\pm$0.03} & 61.5 & \spos{9.0}{\tiny$\pm$0.05} & 64.0 & \spos{6.7}{\tiny$\pm$0.17} & 55.9 & \spos{13.4}{\tiny$\pm$0.29} & 68.4 & \spos{4.2}{\tiny$\pm$0.12} \\
OxfordPets & 95.6 & 94.5 & \spos{0.4}{\tiny$\pm$0.18} & 94.6 & \spos{0.2}{\tiny$\pm$0.22} & 95.9 & \spos{0.2}{\tiny$\pm$0.02} & 96.0 & \spos{0.3}{\tiny$\pm$0.17} & 95.2 & \spos{0.7}{\tiny$\pm$0.04} & 96.3 & \szero{0.0}{\tiny$\pm$0.13} \\
STL10 & 99.5 & 98.6 & \spos{0.5}{\tiny$\pm$0.01} & 98.7 & \spos{0.2}{\tiny$\pm$0.05} & 99.5 & \szero{0.0}{\tiny$\pm$0.01} & 99.3 & \spos{0.2}{\tiny$\pm$0.01} & 98.4 & \spos{1.0}{\tiny$\pm$0.01} & 99.5 & \szero{0.0}{\tiny$\pm$0.02} \\
CIFAR100 & 93.3 & 80.2 & \spos{3.4}{\tiny$\pm$0.09} & 80.9 & \spos{3.4}{\tiny$\pm$0.18} & 86.5 & \spos{1.8}{\tiny$\pm$0.03} & 82.4 & \spos{3.9}{\tiny$\pm$0.15} & 77.3 & \spos{8.0}{\tiny$\pm$0.09} & 85.6 & \spos{1.6}{\tiny$\pm$0.18} \\
CIFAR10 & 99.2 & 97.1 & \spos{0.6}{\tiny$\pm$0.05} & 97.2 & \spos{0.5}{\tiny$\pm$0.06} & 98.5 & \spos{0.3}{\tiny$\pm$0.03} & 97.9 & \spos{0.7}{\tiny$\pm$0.05} & 96.6 & \spos{1.7}{\tiny$\pm$0.02} & 98.4 & \spos{0.3}{\tiny$\pm$0.05} \\
Food101 & 95.5 & 88.8 & \spos{3.9}{\tiny$\pm$0.05} & 89.9 & \spos{3.0}{\tiny$\pm$0.10} & 93.4 & \spos{1.1}{\tiny$\pm$0.01} & 92.2 & \spos{2.3}{\tiny$\pm$0.03} & 87.8 & \spos{6.0}{\tiny$\pm$0.06} & 94.2 & \spos{0.5}{\tiny$\pm$0.03} \\
FashionMNIST & 95.8 & 81.7 & \spos{2.0}{\tiny$\pm$0.33} & 82.5 & \spos{1.9}{\tiny$\pm$0.25} & 91.8 & \spos{0.7}{\tiny$\pm$0.05} & 91.0 & \spos{0.8}{\tiny$\pm$0.10} & 88.8 & \spos{2.5}{\tiny$\pm$0.15} & 90.7 & \spos{0.7}{\tiny$\pm$0.09} \\
EMNIST & 99.8 & 62.1 & \spos{23.9}{\tiny$\pm$0.61} & 75.1 & \spos{16.6}{\tiny$\pm$0.43} & 96.7 & \spos{1.3}{\tiny$\pm$0.13} & 99.3 & \sneg{--0.1}{\tiny$\pm$0.03} & 98.9 & \spos{0.2}{\tiny$\pm$0.04} & 99.1 & \sneg{--0.1}{\tiny$\pm$0.05} \\
KMNIST & 98.8 & 21.3 & \spos{21.2}{\tiny$\pm$1.22} & 28.9 & \spos{20.6}{\tiny$\pm$1.22} & 86.5 & \spos{5.8}{\tiny$\pm$0.15} & 77.7 & \spos{10.3}{\tiny$\pm$0.11} & 62.2 & \spos{23.9}{\tiny$\pm$0.03} & 75.7 & \spos{11.5}{\tiny$\pm$0.21} \\
RenderedSST2 & 85.4 & 65.5 & \spos{12.2}{\tiny$\pm$0.34} & 67.6 & \spos{11.1}{\tiny$\pm$0.36} & 82.5 & \spos{3.0}{\tiny$\pm$0.11} & 82.8 & \spos{2.5}{\tiny$\pm$0.00} & 81.5 & \spos{5.0}{\tiny$\pm$0.24} & 86.1 & \sneg{--0.6}{\tiny$\pm$0.28} \\
\midrule
\rowcolor{grayrow}\textbf{Avg.} & 94.0 & 73.8 & \spos{6.4}{\tiny$\pm$0.03} & 75.5 & \spos{5.5}{\tiny$\pm$0.04} & 89.3 & \spos{1.5}{\tiny$\pm$0.06} & 87.2 & \spos{2.6}{\tiny$\pm$0.04} & 82.6 & \spos{6.6}{\tiny$\pm$0.06} & 89.1 & \spos{1.2}{\tiny$\pm$0.04} \\
\bottomrule
\end{tabular}}
\end{table}

%% file: figs/supp/fig_analysis_remaining.tex
\begin{figure}[t]
\centering
\includegraphics[width=\linewidth]{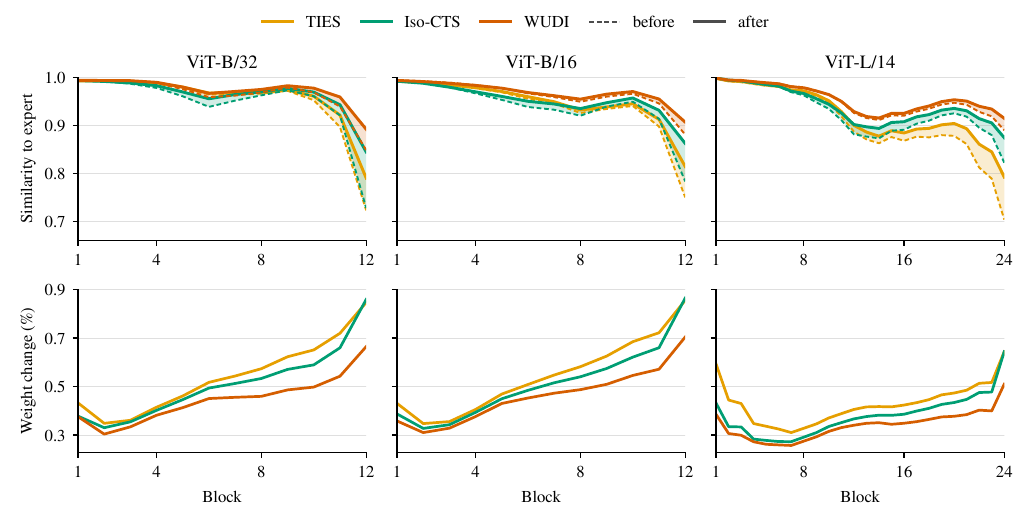}
\caption{Top: cosine similarity between the class-token representations of the merged model and the corresponding expert at the output of each block, on real test images, before (dashed) and after (solid) applying Merge++. The shaded region is the gain. Bottom: relative change of the weights of each block in percent, $\|\Delta W\|/\|W\|$, where $W$ are the weights of the base merge and $\Delta W$ the change made by Merge++. Refinement changes every block by less than 1\% of its weights, yet closes 22--43\% of the gap in similarity to the experts at the last block. It thus makes small, targeted corrections rather than retraining the network. Shown for TIES, Iso-CTS and WUDI with 8 tasks.}
\label{fig:supp_analysis}
\end{figure}